%% file: TNNLS.tex
\documentclass[lettersize,journal]{IEEEtran}
\usepackage{amsmath,amsfonts}
\usepackage{amsmath}
\usepackage{algorithm}
\usepackage{algpseudocode}  % algpseudocode replaces the standard algorithmic package
\usepackage{array}
\usepackage{booktabs}
\usepackage{adjustbox}
\usepackage{amsmath}
\usepackage{subcaption}
\usepackage{algorithm}
\usepackage{algpseudocode}
\usepackage{textcomp}
\usepackage{stfloats}
\usepackage{url}
\usepackage{tcolorbox}
\usepackage{verbatim}
\usepackage{graphicx}
\usepackage{multirow}
\usepackage{threeparttable}
\usepackage[numbers]{natbib}
\usepackage{booktabs} 
\usepackage{amssymb}% http://ctan.org/pkg/amssymb
\usepackage{pifont}% http://ctan.org/pkg/pifont
\begin{document}

\title{Toward Understanding BERT-Like Pre-Training for DNA Foundation Models}

\author{Chaoqi Liang, Lifeng Qiao, Peng Ye, Nanqing Dong, Jianle Sun, Weiqiang Bai, Yuchen Ren, Xinzhu Ma, Hongliang Yan, Chunfeng Song~\IEEEmembership{Senior Member, IEEE}, Wanli Ouyang~\IEEEmembership{Senior Member, IEEE}, Wangmeng Zuo~\IEEEmembership{Senior Member, IEEE}
\thanks{IEEE is the copyright holder. The manuscript is currently under review.}
\thanks{The first three authors contributed equally to this work.}
\thanks{C. Liang is with the School of Computer Science and Technology, Harbin Institute of Technology, Harbin 150001, China; and also with Shanghai Artificial Intelligence Laboratory, Shanghai 200232, China. (email: liangchaoqi@pjlab.org.cn)}
\thanks{
L. Qiao, P. Ye, N. Dong, J. Sun, W. Bai, Y. Ren, X. Ma, C. Song, and W. Ouyang are with Shanghai Artificial Intelligence Laboratory, Shanghai 200232, China. (email: dongnanqing@pjlab.org.cn)}
\thanks{H. Yan is with Guangzhou National Laboratory, Guangzhou 510200, China. (email: yan\_hongliang@gzlab.ac.cn)}
\thanks{W. Zuo is with the School of Computer Science and Technology, Harbin Institute of Technology, Harbin 150001, China. (email: wmzuo@hit.edu.cn)}
}
\markboth{Journal of \LaTeX\ Class Files,~Vol.~14, No.~8, August~2021}%
{Shell \MakeLowercase{\textit{et al.}}: A Sample Article Using IEEEtran.cls for IEEE Journals}

% \author{Anonymous TNNLS submission
% \thanks{Anonymous Author(s)}}

%\markboth{Journal of \LaTeX\ Class Files,~Vol.~18, No.~9, September~2020}%
%{How to Use the IEEEtran \LaTeX \ Templates}


\maketitle

\begin{abstract}
With the success of large-scale pre-training in language tasks, there is an increasing trend of applying it to the domain of life sciences. In particular, pre-training methods based on DNA sequences have received increasing attention because of their potential to capture general information about genes. However, existing pre-training methods for DNA sequences largely rely on direct adoptions of BERT pre-training from NLP, lacking a comprehensive understanding and a specifically tailored approach. To address this research gap, we provide the first empirical study with three insightful observations. Based on the empirical study, we notice that overlapping tokenizer can benefit the fine-tuning of downstream tasks but leads to inadequate pre-training with fast convergence. 
% : 1) In the fine-tuning phase of downstream tasks, when using K-mer overlapping tokenizer instead of K-mer non-overlapping tokenizer, both overlapping and non-overlapping pre-training weights show consistent performance improvement. 2) Using the K-mer overlapping tokenizer quickly produces clear K-mer tokens' embedding space during pre-training. It quickly reduces the pre-training loss to a very low level, while using non-overlapping K-mer tokenizer results in less distinct K-mer tokens' embedding space and continuously decreases the pre-training loss. 3) Using overlapping tokenizer causes the self-attention in the intermediate layers of pre-trained models to overly focus on certain tokens, reflecting that these layers are not adequately optimized. 
% In summary, overlapping tokenizer can benefit the fine-tuning of downstream tasks but leads to inadequate pre-training with fast convergence. 
To unleash the pre-training potential, we introduce a novel approach called RandomMask, which gradually increases the task difficulty of BERT-like pre-training by continuously expanding its mask boundary, forcing the model to learn more knowledge. RandomMask is simple but effective, achieving state-of-the-art performance across 6 downstream tasks. RandomMask achieves a staggering 68.16\% in Matthew's correlation coefficient for Epigenetic Mark Prediction, a groundbreaking increase of 19.85\% over the baseline and a remarkable 3.69\% improvement over the previous state-of-the-art result.

\end{abstract}

\begin{IEEEkeywords}
Large-scale Pre-Training, Tokenizer, Masked Language Modeling, DNA.
\end{IEEEkeywords}

\input{01Introduce}

\input{02RelatedWork}
\input{03Exploratory_Experiments_and_Observations}

\input{04Method}
\input{05Experiments}

\input{06Ablations}
\input{07Conclusions}

\bibliographystyle{IEEEtran}
\bibliography{TNNLS}
\input{Appendix}
% \newpage
 
\end{document}

%% file: 01Introduce.tex
\section{Introduction}

\IEEEPARstart{I}{n} recent years, the integration of Transformer architectures, extensive datasets, and self-supervised pre-training techniques has significantly advanced the field of natural language processing (NLP) \cite{devlin2018bert, floridi2020gpt, zhou2021topicbert, zhao2023survey, hua2023improving}.  Similarly, these advances find an echo in the study of DNA sequences, where complex interactions among elements such as promoters, enhancers, and transcription factor binding sites mirror the intricate semantic relationships in language \cite{khoury1983enhancer, riethoven2010regulatory, guo2022context, wang2024grace}. The power of pre-trained language models in distinguishing these subtle and interconnected patterns springs from pre-training on extensive, unlabeled data. Fortunately, projects like the Human Genome Project have provided a wealth of DNA sequence data \cite{gibbs2020human}, setting the stage for developing genomic pre-training models.

The prospect of utilizing pre-trained language models to uncover the hidden knowledge from vast DNA sequences is highly promising. Pioneering models like DNABERT \cite{ji2021dnabert}, LOGO \cite{yang2022integrating}, and the Nucleotide Transformer \cite{dalla2023nucleotide} have demonstrated significant progress in the analysis of DNA sequences by BERT-like pre-training model. Considering that current DNA modeling primarily focuses on understanding existing sequences rather than generating new ones, BERT-like models' bidirectional context understanding capability is typically more crucial than the unidirectional generative capability of GPT-like models.

Significant advancements have been made in DNA foundation models recently, influenced by the success of BERT. DNABERT, introduced by~\cite {ji2021dnabert}, applies BERT-like architectures to learn representations of DNA sequences. By leveraging Transformers' bidirectional nature, DNABERT captures dependencies and relationships between nucleotides, enabling a deeper understanding of genetic information~\citep{le2021transformer}. It has demonstrated enhanced performance on tasks like DNA sequence classification, variant calling, and gene expression prediction. Another notable advancement is the Nucleotide Transformer (NT) proposed by~\cite {dalla2023nucleotide}. NT utilizes a significantly larger number of parameters compared to DNABERT, leading to notable performance enhancements. As the field continues to evolve, further refinements and novel approaches are expected, leading to more advanced analysis and interpretation of genetic information~\cite {nguyen2023hyenadna, zhou2023dnabert}.

\begin{figure*}[t]
  \centering
  \begin{minipage}[t]{.33\linewidth}
    \centering
    \includegraphics[width=\linewidth]{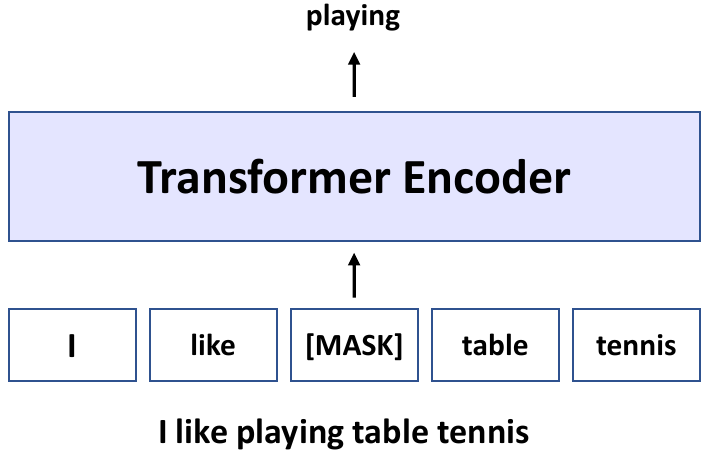}
    % \vskip -0.07in
    \subcaption{MLMs for NLP.}
  \end{minipage}%
  \hfill
  \begin{minipage}[t]{.33\linewidth}
    \centering
    \includegraphics[width=\linewidth]{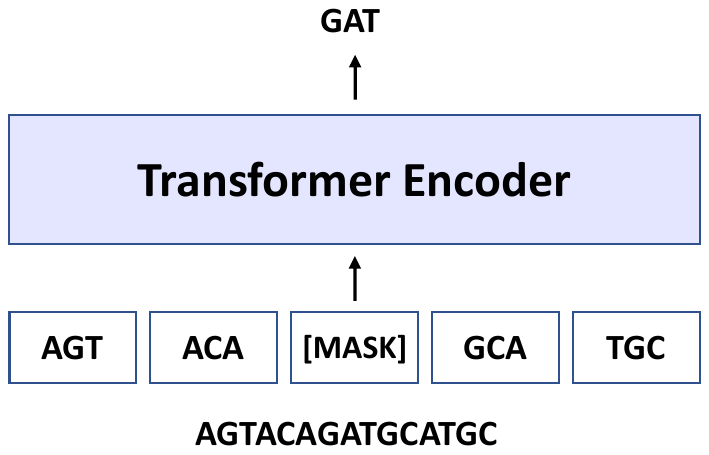}
    % \vskip -0.07in
    \subcaption{MLMs for DNA non-overlapping 3-mer tokenizer.}
  \end{minipage}
  \hfill
  \begin{minipage}[t]{.33\linewidth}
    \centering
    \includegraphics[width=\linewidth]{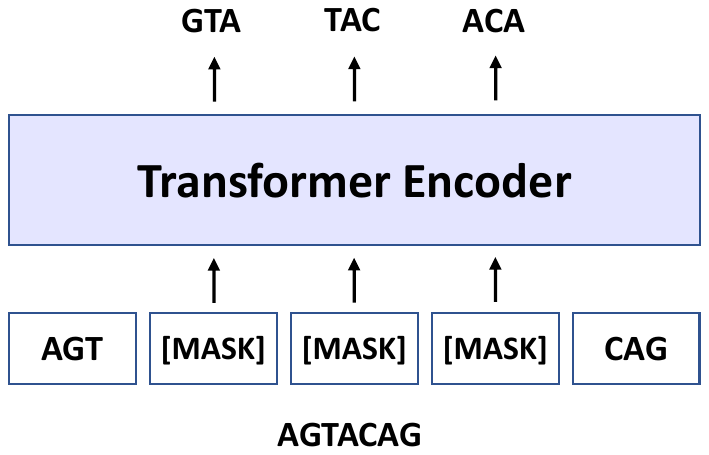}
    % \vskip -0.07in
    \subcaption{MLMs for DNA overlapping 3-mer tokenizer.}
  \end{minipage}
  \hfill
  % \vskip -0.15in
  \caption{Comparison of MLMs: From NLP to DNA sequence analysis. In the experiments of this paper, both DNABERT and NT utilized 6-mer. For illustrative purposes, the figures use 3-mer as a representation.}
  \label{figure_struct}
\end{figure*}

\begin{figure*}[t]
  \centering
    \includegraphics[width=\linewidth]{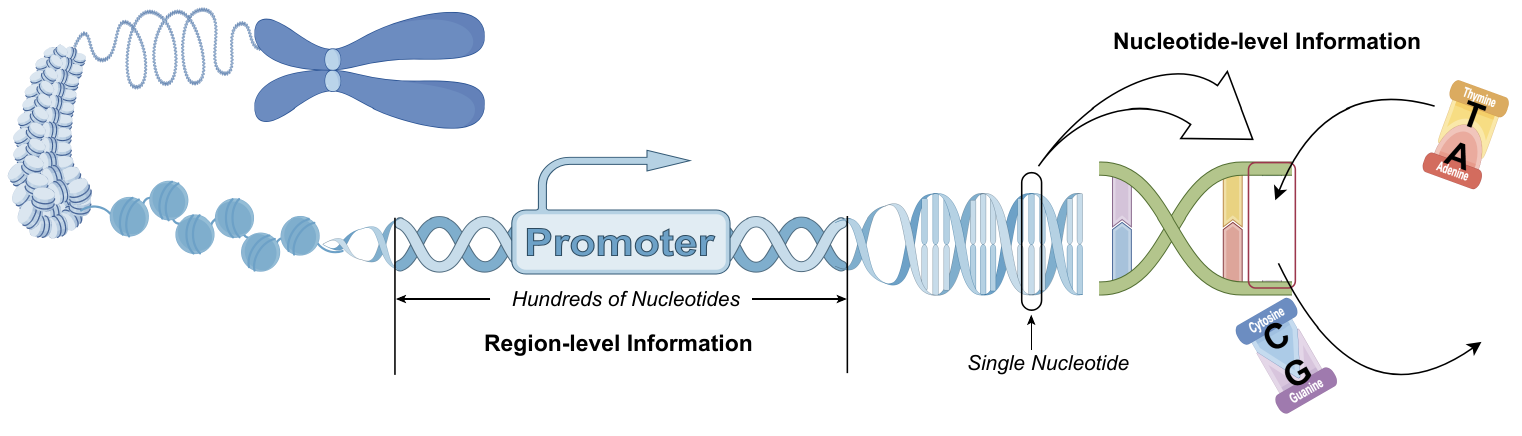}
  \label{figure_2}
  \vspace{-4mm}
  \caption{{Illustration of the region- and nucleotide-resolution information for DNA sequence modeling. DNA modeling requires the capture of information at two distinct levels. At the regional level, patterns of functional elements in DNA sequences span tens to hundreds of nucleotides, such as promoters and enhancers, which act as integrated units to regulate gene expressions. Besides, capturing information at the nucleotide resolution is also crucial, as variations in a single nucleotide of DNA sequences can result in significant alterations to gene functions.}}
  \label{bio_fig}
\end{figure*}

However, pre-trained models for DNA sequences often directly leverage NLP methods such as BERT \cite{devlin2018bert}, neglecting the unique characteristics of DNA sequences. Figure~\ref{figure_struct} illustrates both overlapping and non-overlapping tokenizer strategies employed in DNA analysis, such as DNABERT and Nucleotide Transformer (NT) \cite{ dalla2023nucleotide}. Despite the sophisticated tokenizer strategies, these models usually fail to capture the characteristics of DNA sequences, as shown in Figure~\ref{bio_fig}. First, genomes contain functional elements with specific long sequence patterns ranging from tens to hundreds of long nucleotides, such as promoters (\cite{moore2013dna, oubounyt2019deepromoter}),on building up \emph{region-level} genomic information. Furthermore, as exemplified by the simple genetic substitution (\emph{e.g.}~GAA to GTA) that leads to sickle cell anemia \cite{kato2018sickle}, even a single nucleotide change in the genome can deeply affect gene function, making capture of the \emph{nucleotide-level} information crucial as well. This complexity underscores the necessity for models tailored to DNA sequences' region-level and nucleotide-level information. 
% these models fail to fully account for the critical impact of minor nucleotide variations, such as the simple genetic substitution (\emph{e.g.}~GAA to GTA) that leads to sickle cell anemia \cite{kato2018sickle}. Furthermore, DNA tasks vary greatly from identifying biomolecular interactions in Epigenetic Marks Prediction \cite{moore2013dna} to examining extensive sequences in Promoter Detection \cite{oubounyt2019deepromoter}, as shown in Figure~\ref{bio_fig}.
% \footnote{More details are summarized in Table~\ref{tabel_1} in Appendix~\ref{DNA_Downstream_Tasks}.} 
% This diversity underscores the necessity for models tailored to DNA sequences' specific properties and nuances. 

A deeper understanding of BERT-like models for DNA is needed to develop pre-training methods suitable to the DNA characteristics. Specifically, our observations reveal several crucial phenomena:  1) Regardless of the source of pre-trained weights—whether from models using overlapping or non-overlapping tokenizer, using overlapping tokenizer consistently improves performance in downstream tasks. This improvement is likely due to its sensitivity to single nucleotide changes. 2) During pre-training, overlapping tokenizer rapidly produces distinct K-mer embeddings and achieves exceptionally low losses, whereas non-overlapping tokenizer tends to produce more ambiguous embeddings and continuous loss reduction. 3) Models pre-trained with overlapping tokenizer tend to show a pattern in their intermediate layers, concentrating self-attention narrowly on specific tokens. It may suggest an issue of under-training in these layers, and the model's ability to model regional-level information is insufficient \cite{jawahar2019does}. In summary, while the overlapping tokenizer method improves fine-tuning performance, it also faces challenges during pre-training, including rapid convergence and potential under-training risk.

Building upon these insights, we believe that modeling DNA sequences should consider single nucleotide features and region-level information. We propose RandomMask, a technique that increases the complexity of pre-training tasks for models using overlapping tokenizer. The overlapping tokenizer helps the model capture DNA single nucleotide features, and RandomMask lets the model learn DNA region-level information by reconstructing DNA sequences of different lengths. RandomMask dynamically expands masking boundaries during BERT-like pre-training, introducing evolving challenges. Observing the mechanism of attention in the middle layer effectively addresses the issue of rapid convergence observed in these models, which can otherwise lead to a superficial understanding of complex DNA patterns. 
%Meanwhile, RandomMask maintains superior performance in downstream tasks that require precise nucleotide resolution. 

Empirically, RandomMask has set new benchmarks, achieving state-of-the-art 
(SOTA) performance on 6 downstream tasks \cite{zhou2023dnabert, de2022deepstarr}. In the task of epigenetic mark prediction, RandomMask achieved a mean Matthew’s correlation coefficient of 68.16\%, improving the baseline by 19.85\% and exceeding the previous SOTA by 3.69\%.

The contributions of this paper are summarized as follows:

\begin{itemize}
\item 
We conducted a thorough analysis of BERT-like pre-training for DNA sequences. Our findings reveal that the K-mer overlapping tokenizer enhances performance during the fine-tuning phase, regardless of whether models are pre-trained with overlapping or non-overlapping weights. However, the common overlapping tokenizer method leads to rapid convergence and under-training during the pre-training phase.
\item
To address these issues and unleash the potential of pre-training, we introduced RandomMask. This novel method dynamically expands the masking boundaries, increasing the complexity of the pre-training task and encouraging the model to learn richer and more robust knowledge of DNA sequences.
\item 
We evaluated RandomMask on 6 downstream tasks, where it consistently achieved superior performance. Notably, in the epigenetic mark prediction task, RandomMask reached a mean Matthew’s correlation coefficient of 68.16\%, surpassing the baseline by 19.85\% and exceeding the current SOTA by 3.69\%.
\end{itemize}

%% file: 02RelatedWork.tex
\section{Preliminaries}
\subsection{K-mer tokenizer}

K-mer tokenizer involves dividing DNA sequences into subsequences of length K using a sliding window mechanism. Here, ``K'' represents the window size and determines the length of each subsequence. This framework has two commonly used strategies: \textbf{Overlapping} and \textbf{Non-overlapping} tokenizer. Overlapping tokenizer, used by DNABERT, involves a window size of \( K \) and a stride of 1. This approach would tokenize the DNA sequence ``ATGACG'' into subsequences ATG, TGA, GAC, and ACG using a 3-mer window. In contrast, non-overlapping tokenizer, employed by the Nucleotide Transformer, uses both a window size and stride of \( K \). This results in subsequences like ATG and ACG for the same sequence using a 3-mer window.

\subsection{Significance of Single Nucleotide Resolution}

Single nucleotide resolution is crucial for a wide range of DNA-related tasks. Recognizing its significance, Nguyen et al. emphasized this aspect in their study HyenaDNA~\citep{nguyen2023hyenadna}. They argued that a stride of 1 is essential for models to identify and extract detailed information about individual nucleotides accurately. From this perspective, they advocated for a single nucleotide tokenizer strategy that employs a stride of 1 to achieve enhanced resolution at the single nucleotide level.

%% file: 03Exploratory_Experiments_and_Observations.tex
% \section{Exploratory Attempts at Overlapping and Non-overlapping tokenizer}
\section{Observations}

To examine the effect of different tokenizer methods, we performed two exploratory experiments and gained three insightful observations.

\begin{itemize}
    \item  It is common practice to adopt consistent tokenizer methods for pre-training and fine-tuning. Contrary to this conventional wisdom, which posits that tokenizer inconsistencies may impair the model's ability to apply learned knowledge effectively, our results suggest otherwise.  Overlapping tokenizer consistently outperforms non-overlapping tokenizer in DNA downstream tasks, regardless of the tokenizer method pre-training employed. This finding indicates that overlapping tokenizer is particularly advantageous for DNA sequence analysis by nature.
    \item In order to delve deeper into the underlying differences between overlapping and non-overlapping tokenizer, we conducted an extensive analysis of the pre-training process. This analysis allowed us to gain two more insightful observations: (1) Overlapping tokenizer leads to a more organized embedding space with exceptionally reduced loss, while non-overlapping tokenizer results in a less structured embedding space with a gradual, continuous decrease in loss. (2) The standard MLM task appears insufficiently challenging for models using overlapping tokenizer, thus hindering the sufficient training of attention mechanisms.
\end{itemize}

% \subsection{Some Details on DNA Sequence tokenizer}

\begin{table*}[t]
\centering
\caption{Performance comparison between overlapping and non-overlapping tokenizer in the fine-tuning stage with two pre-trained models pre-trained with different tokenizer methods. It can be seen that the use of overlapping tokenizer in fine-tuning always yields a performance gain, regardless of the type of tokenizer used for pre-training. The results across 6 downstream tasks \cite{zhou2023dnabert} are reported in the metric of MCC. MCC is described in detail in the Subsection \ref{metric}.}
\label{table:downstream_result}
\resizebox{\linewidth}{!}{
\begin{tabular}{@{}lcccccccccl@{}}
\toprule
Model & Pre-training & Fine-tuning & EMP & TF-M & TF-H & PD & CPD & SSP & Avg. \\
\midrule
\multirow{2}{*}{NT \cite{dalla2023nucleotide}} & \multirow{2}{*}{Non-overlapping} & Non-overlapping & {45.37} & 39.81 & 55.25 & {88.43} & 62.56 & 80.39 & 61.97 \\
 &  & Overlapping & \textbf{46.47} & \textbf{61.99} & \textbf{63.95} & \textbf{90.88} & \textbf{68.55} & \textbf{84.34} & \textbf{69.36} \\
\addlinespace
\multirow{2}{*}{DNABERT \cite{ji2021dnabert}} & \multirow{2}{*}{Overlapping} & Non-overlapping & {43.65} & 34.87 & {54.50} & {87.62} & {65.82} & 79.91 & 61.06 \\
 &  & Overlapping & \textbf{51.81} & \textbf{59.60} & \textbf{63.55} & \textbf{90.48} & \textbf{70.47} & \textbf{85.44} & \textbf{70.23} \\
\bottomrule
\end{tabular}
}
\end{table*}
\begin{figure*}[t]
\includegraphics[width=0.98\textwidth]{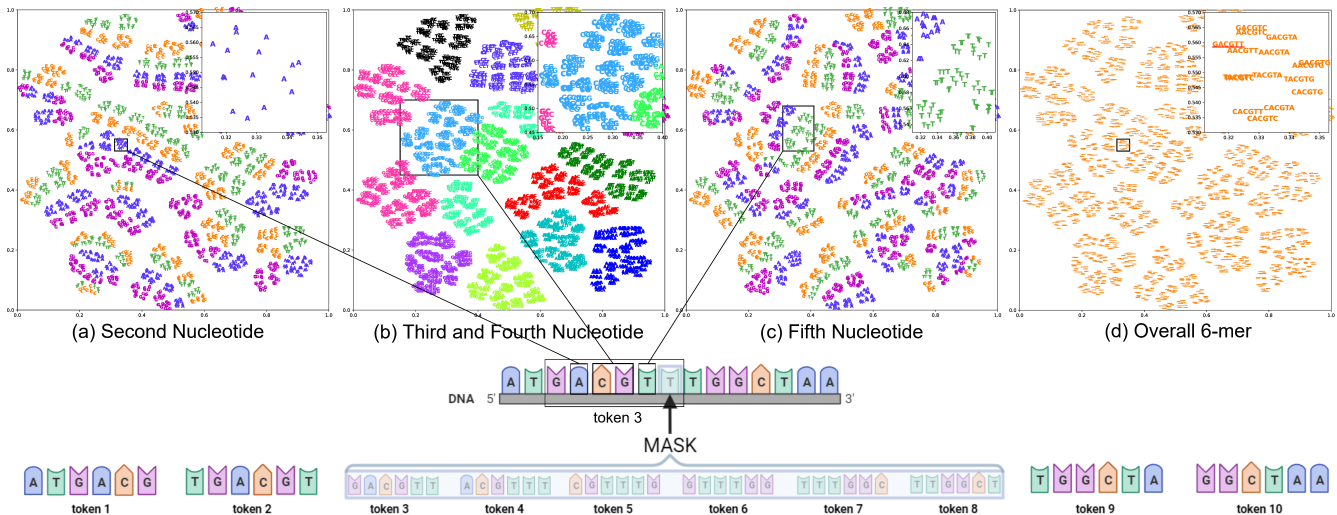}
% \vspace{-1mm}
\caption{Detailed t-SNE visualization of the embedding space learned by DNABERT with overlapping tokenizer. The (a) and (c) plots are the clustering of marginal nucleotides. The (b) plot clusters the two central nucleotides. The (d) plot illustrates the overall 6-mer tokens in the embedding space. The two central nucleotides of a 6-mer token determine the cluster in which it is placed in the embedding space, and the marginal nucleotides determine its placement within the cluster.}
\label{figure:detailed_cluster}
\end{figure*}
\subsection{Fine-tuning Stage}

We performed a series of comparative experiments on diverse downstream benchmark tasks. Two pre-trained models were employed, namely ``DNABERT'' and ``Nucleotide Transformer'', both pre-trained on the whole human genome. ``DNABERT'' was pre-trained using overlapping tokenizer, whereas ``Nucleotide Transformer'' was pre-trained using non-overlapping tokenizer. Then we fine-tuned these two models on the benchmark consisting of 6 downstream tasks\footnote{More details are summarized in Table~\ref{tabel_1} in Subsection~\ref{DNA_Downstream_Tasks}.}. The results are shown in Table~\ref{table:downstream_result}.  

\begin{tcolorbox}[leftrule=1.5mm,top=0.8mm,bottom=0.5mm]
\textbf{Observation 1:}
\begin{itemize}
    \item During the fine-tuning stage, using overlapping tokenizer instead of non-overlapping tokenizer leads to consistent performance improvement for both overlapping (DNABERT) and non-overlapping (Nucleotide Transformer) pre-trained models.
\end{itemize}
\end{tcolorbox}

In Table~\ref{table:downstream_result}, we observe that regardless of the pre-training method employed, models fine-tuned with overlapping tokenizer consistently outperform non-overlapping tokenizer. Specifically, DNABERT demonstrates improvements in all 6 tasks, with an average increase of 9.17\% in MCC. Similarly, the Nucleotide Transformer also improves in all 6 tasks, with an average increase of 7.39\%.

We claim that the performance gap between overlapping and non-overlapping tokenizer stems from the intrinsic superiority of overlapping tokenizer for DNA downstream tasks. Additionally, contrary to conventional belief, which suggests that inconsistency between pre-training and fine-tuning may hinder performance, our finding reveals that directly using overlapping tokenizer leads to a significant improvement in the performance of DNA downstream tasks, regardless of the chosen pre-training method.  

\subsection{Pre-training Stage}

To gain a deeper understanding, we thoroughly analyze the pre-training process. This involves pre-training two models, namely ``DNABERT'' with overlapping tokenizer and ``DNABERT'' non-overlapping tokenizer, on the entire Human Genome~\citep{gibbs2020human}.

\begin{figure*}[t]
  \centering
  \begin{minipage}[t]{.32\linewidth}
    \centering
    \includegraphics[width=\linewidth]{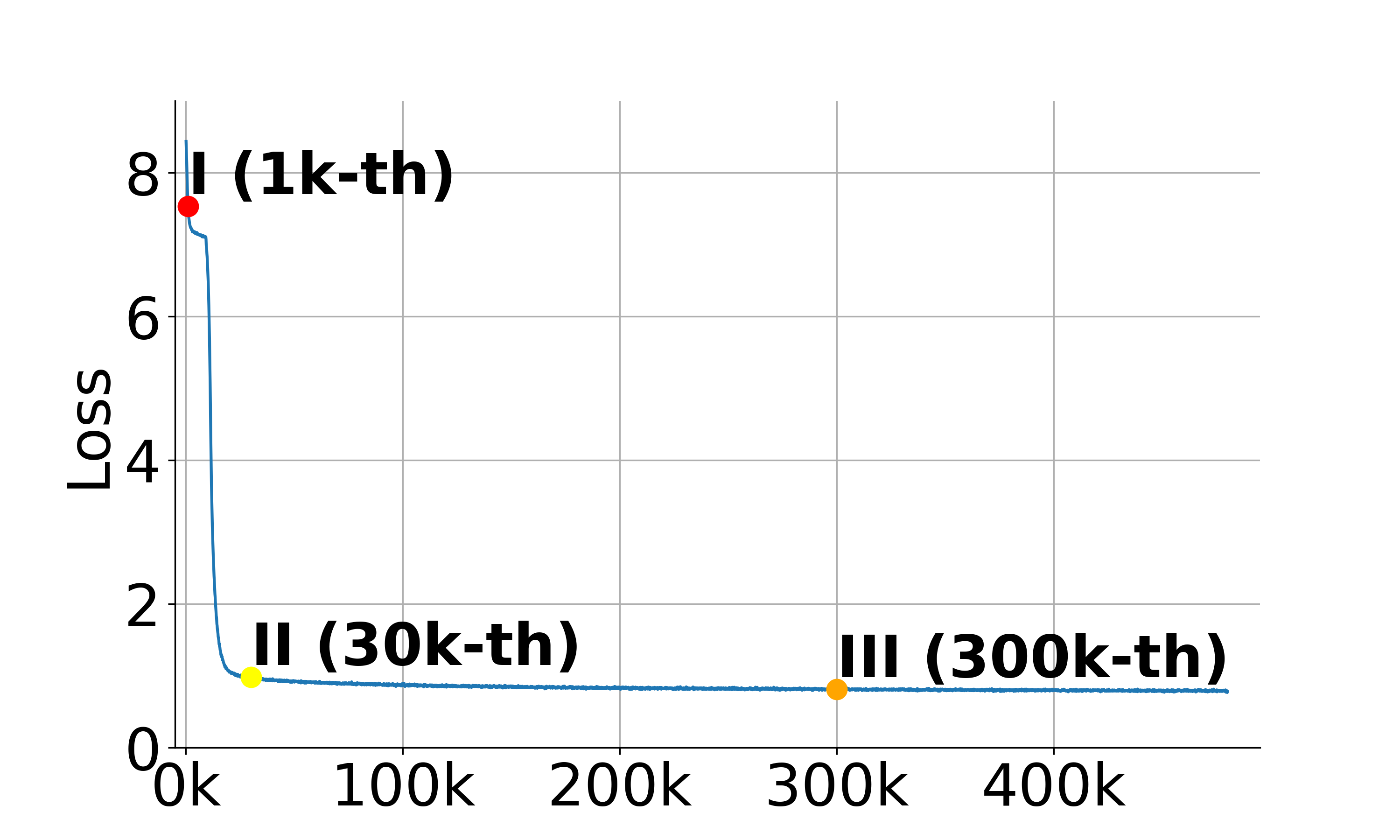}
    \vskip -0.08in
    \subcaption{The loss of Overlapping DNABERT.}
  \end{minipage}%
  \hfill
  \begin{minipage}[t]{.22\linewidth}
    \centering
    \includegraphics[width=\linewidth]{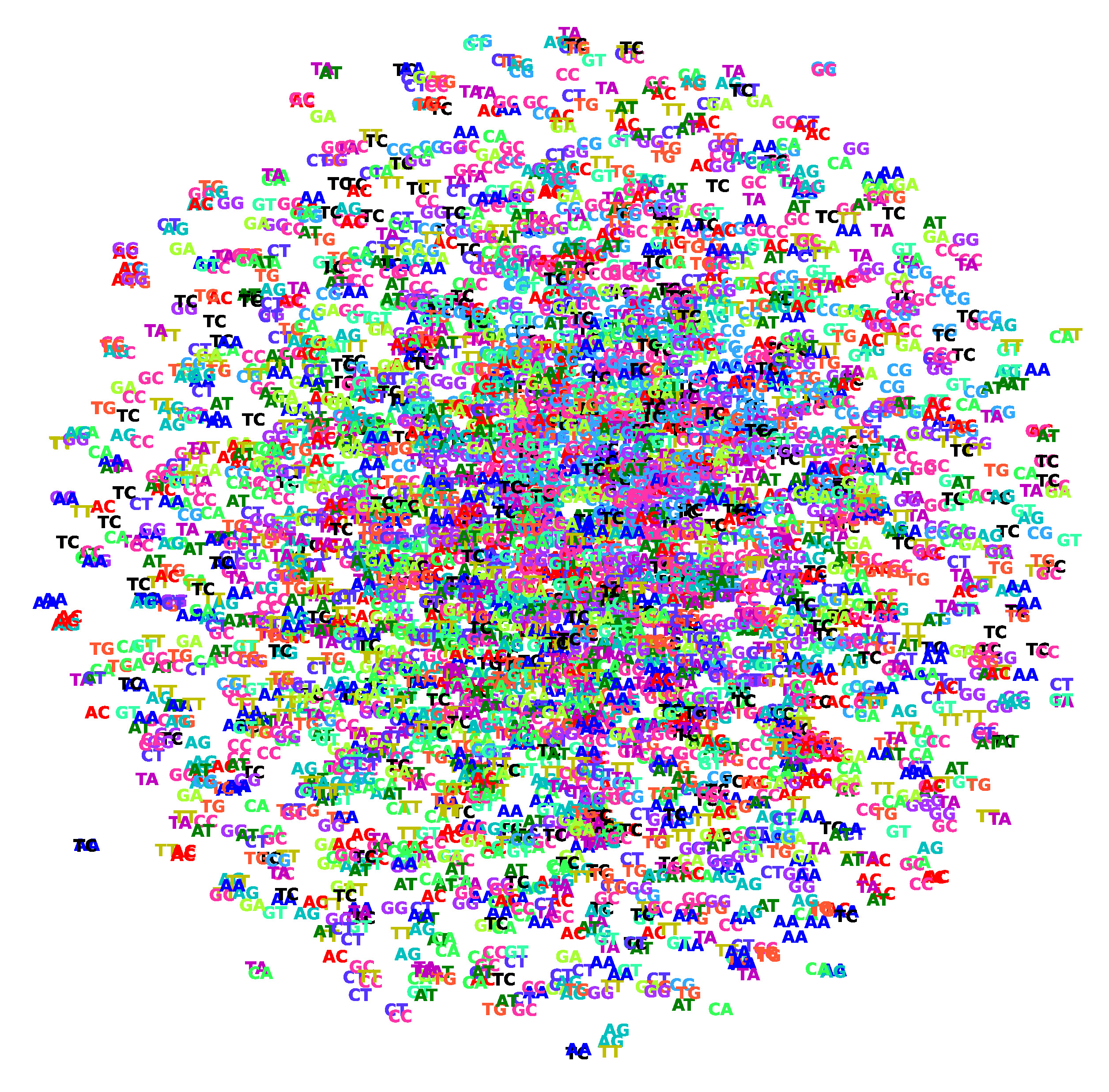}
    \vskip -0.08in
    \subcaption*{I. 1k-th step}
  \end{minipage}
  \hfill
  \begin{minipage}[t]{.22\linewidth}
    \centering
    \includegraphics[width=\linewidth]{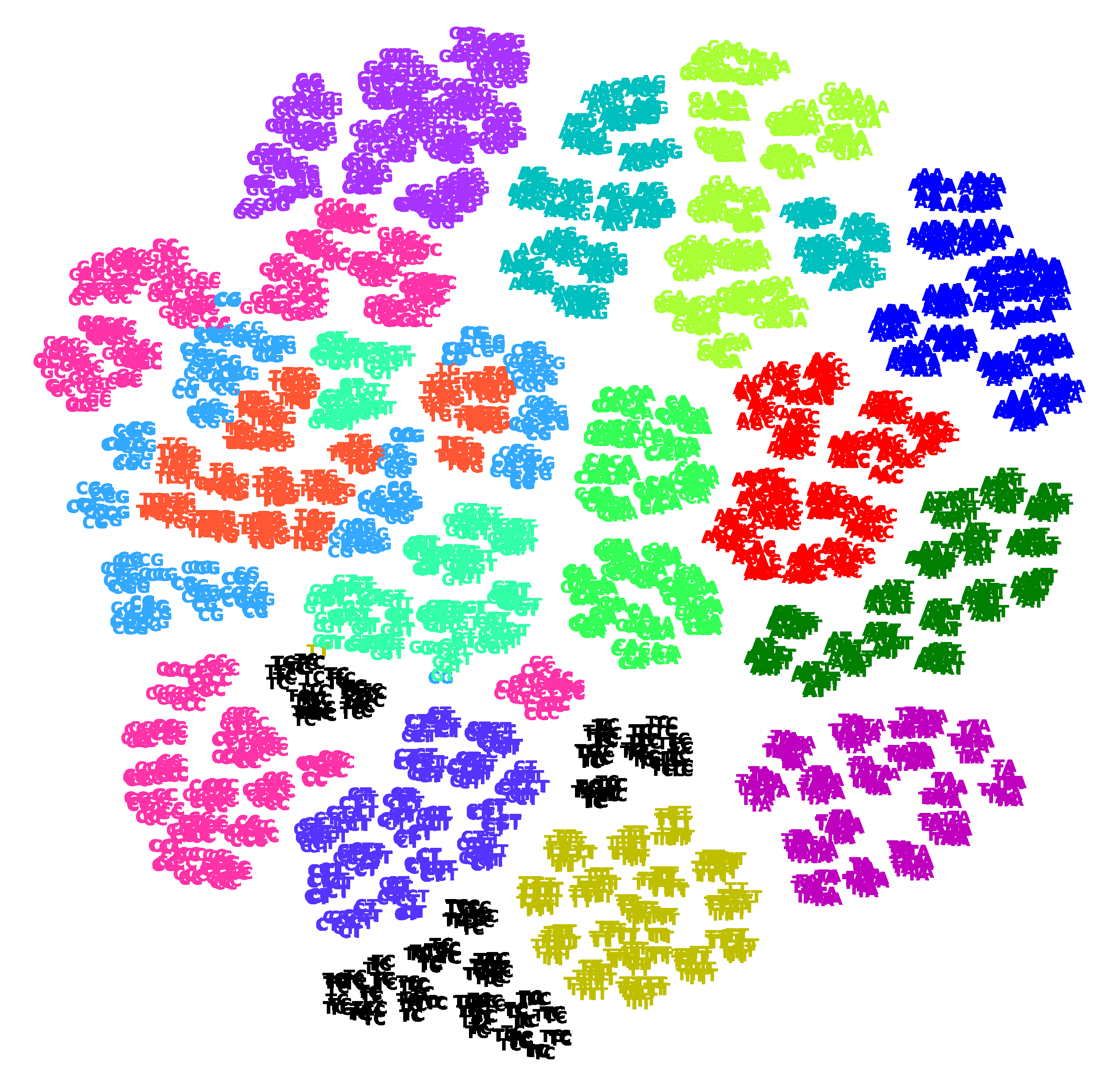}
    \vskip -0.08in
    \subcaption*{II. 30k-th step}
  \end{minipage}
  \hfill
  \begin{minipage}[t]{.22\linewidth}
    \centering
    \includegraphics[width=\linewidth]{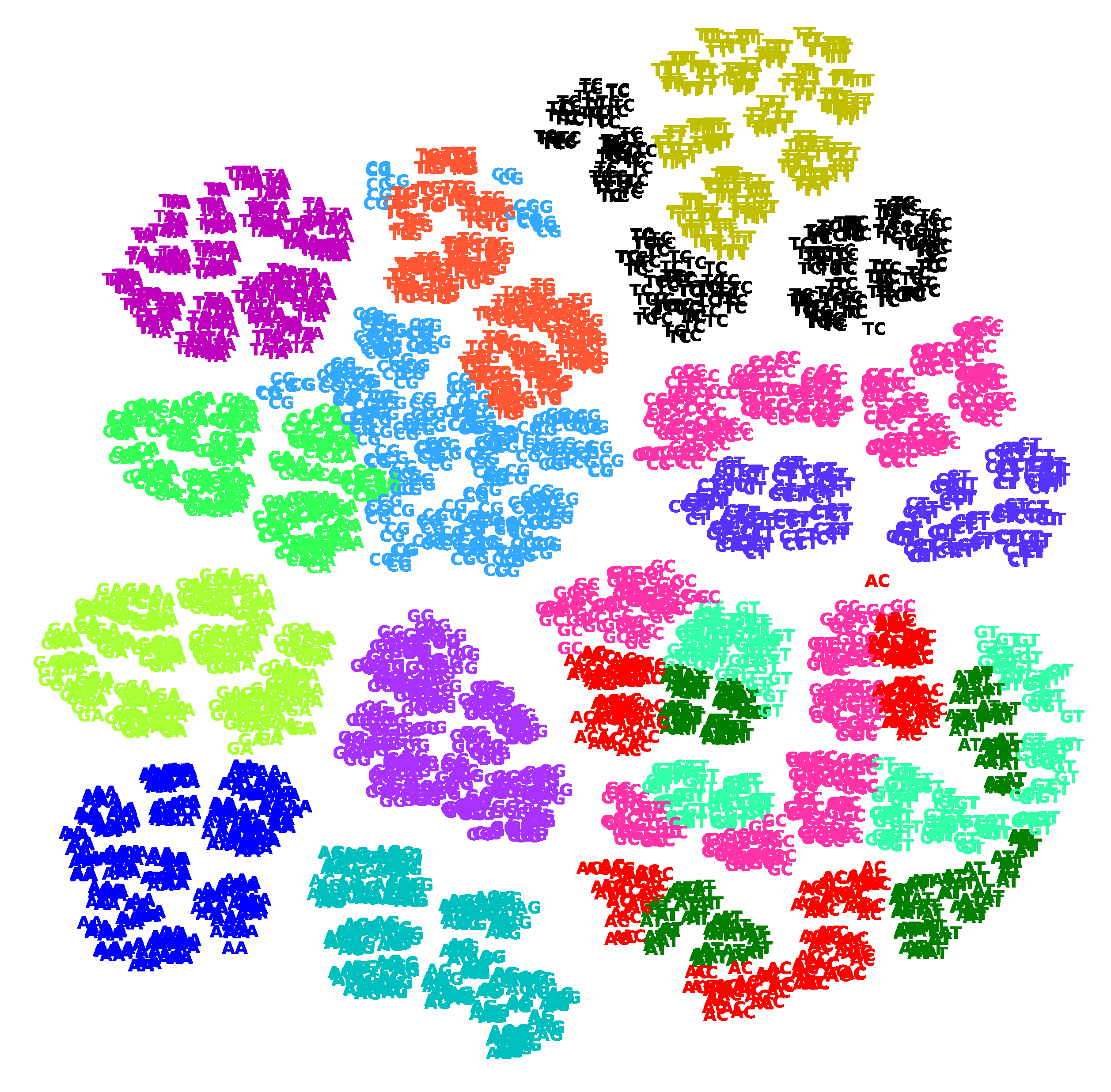}
    \vskip -0.08in
    \subcaption*{III. 300k-th step}
  \end{minipage}
  \hfill
  % Add vertical space between the rows
  \vspace{0.1in} % Adjust this value as needed
  % 第二行的图像
  \begin{minipage}[t]{.32\linewidth}
    \centering
    \includegraphics[width=\linewidth]{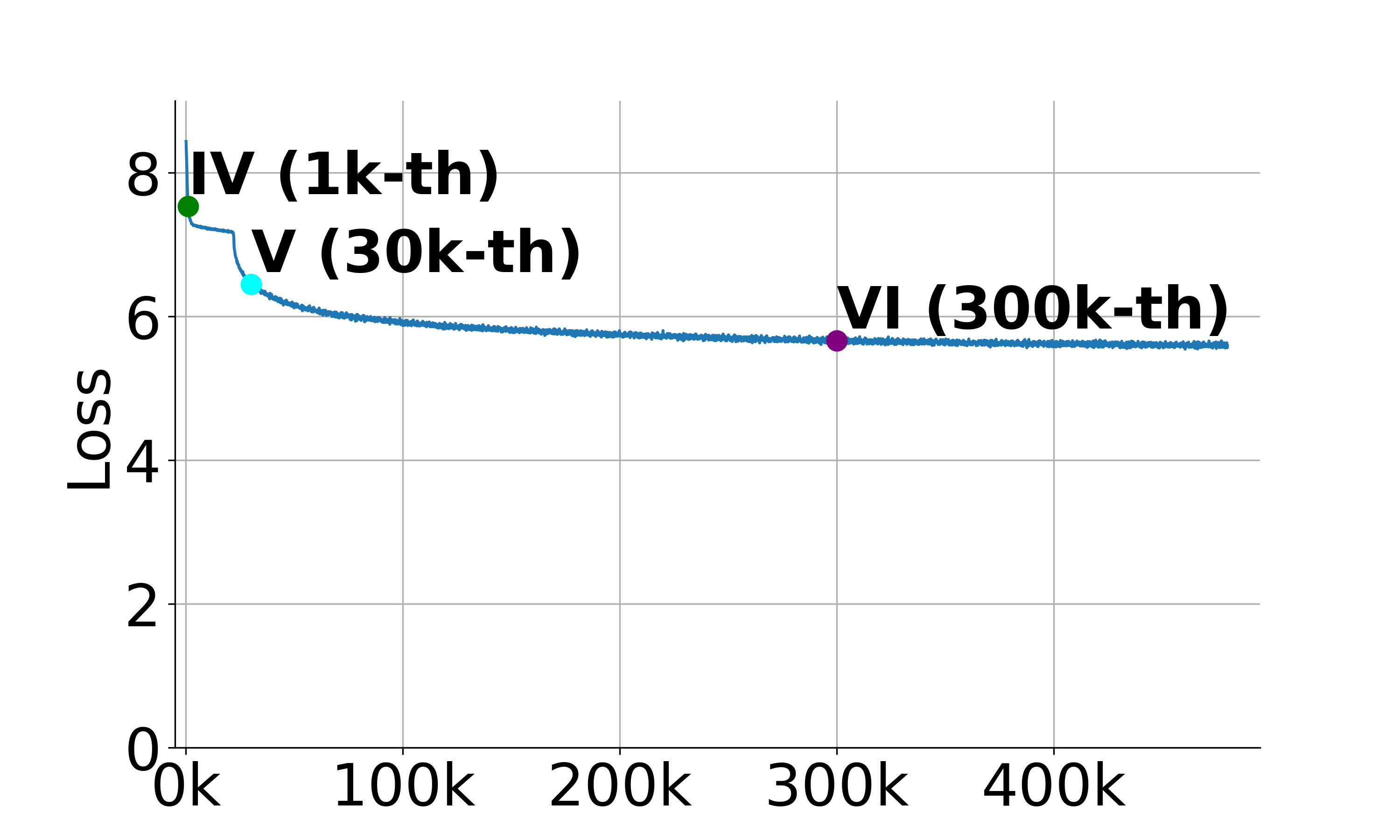}
    \vskip -0.08in
    \subcaption{The loss of Non-Overlapping DNABERT.}
  \end{minipage}%
  \hfill
  \begin{minipage}[t]{.22\linewidth}
    \centering
    \includegraphics[width=\linewidth]{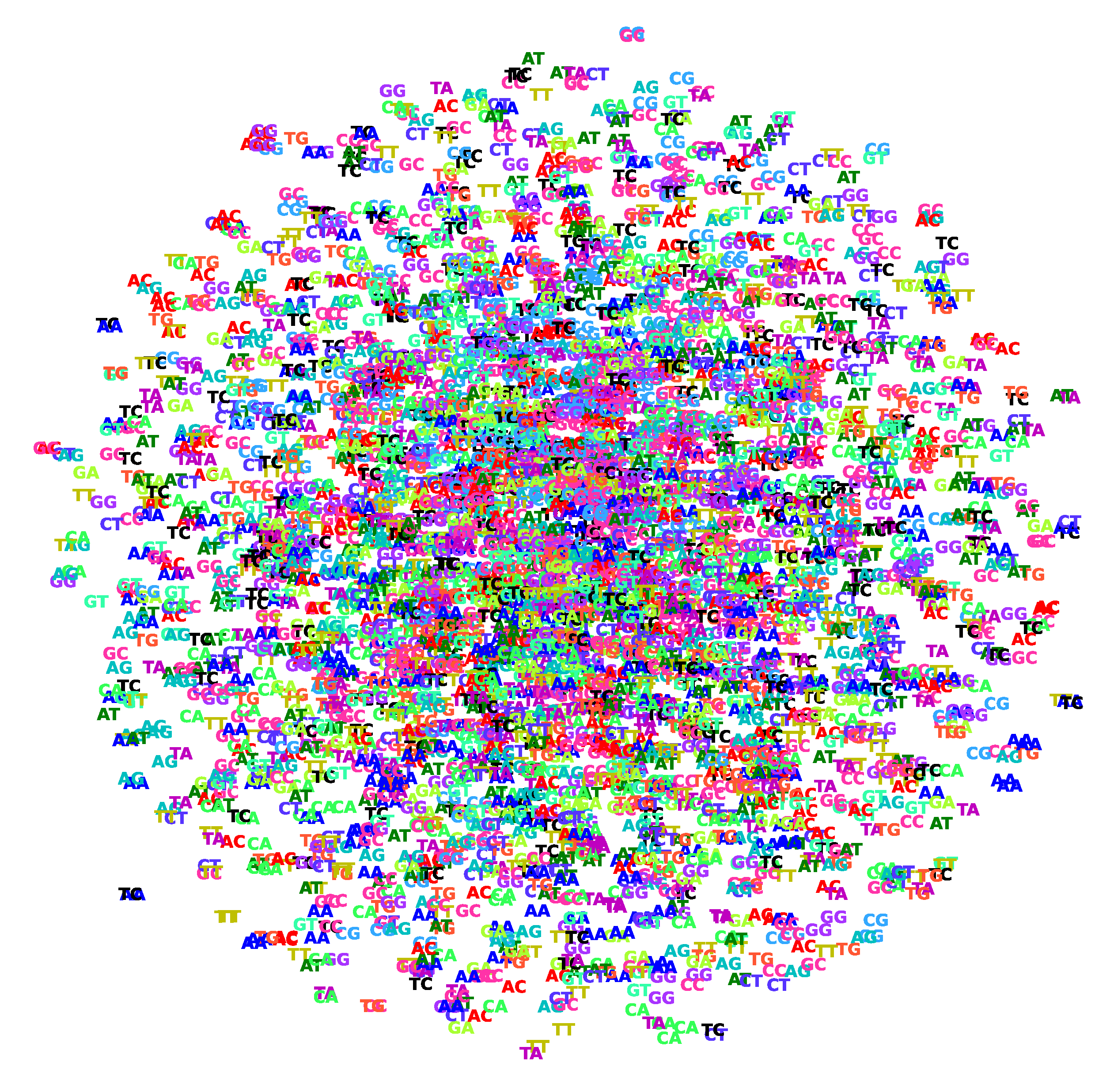}
    \vskip -0.08in
    \subcaption*{IV. 1k-th step}
  \end{minipage}
  \hfill
  \begin{minipage}[t]{.22\linewidth}
    \centering
    \includegraphics[width=\linewidth]{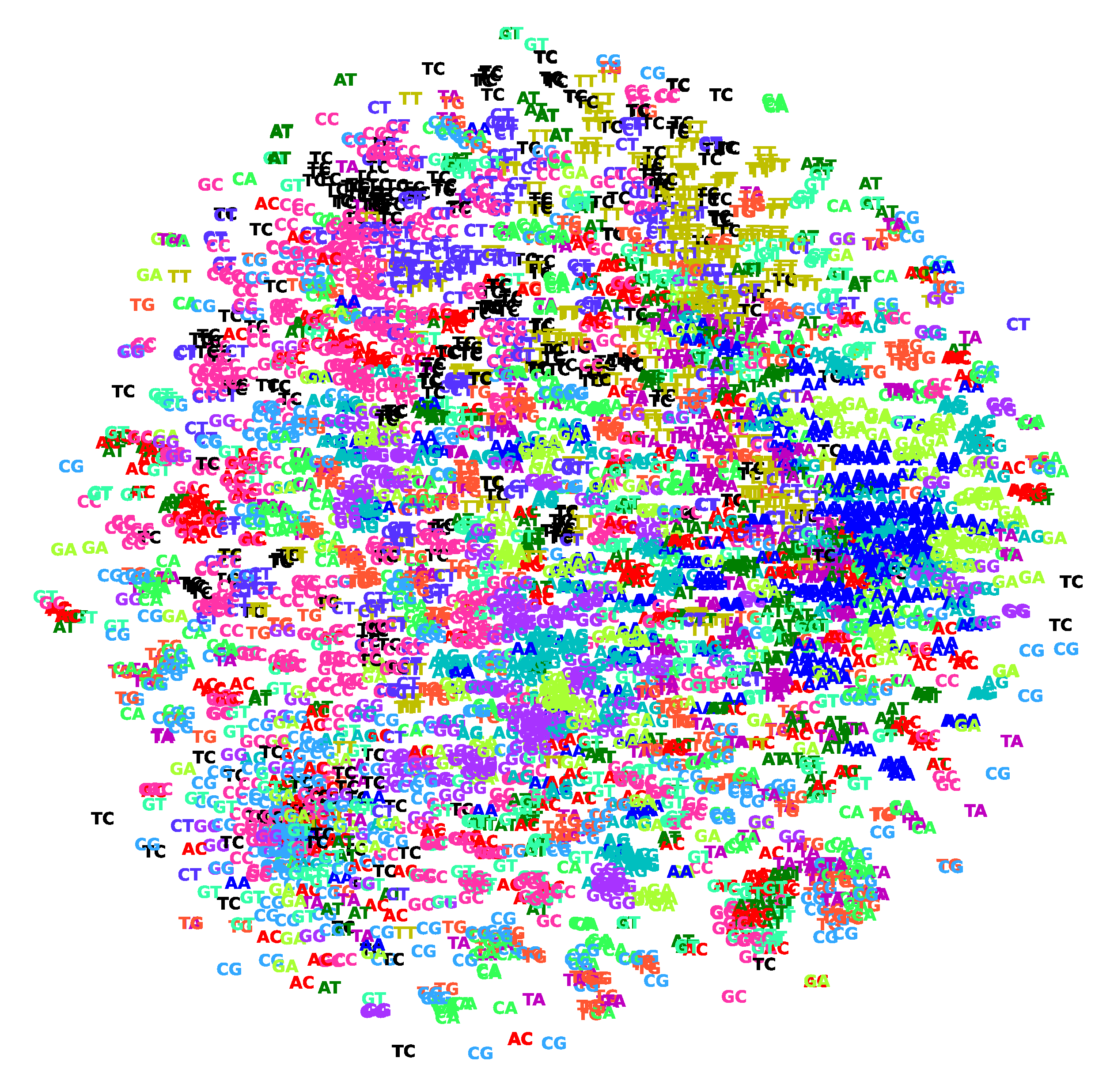}
    \vskip -0.08in
    \subcaption*{V. 30k-th step}
  \end{minipage}
  \hfill
  \begin{minipage}[t]{.22\linewidth}
    \centering
    \includegraphics[width=\linewidth]{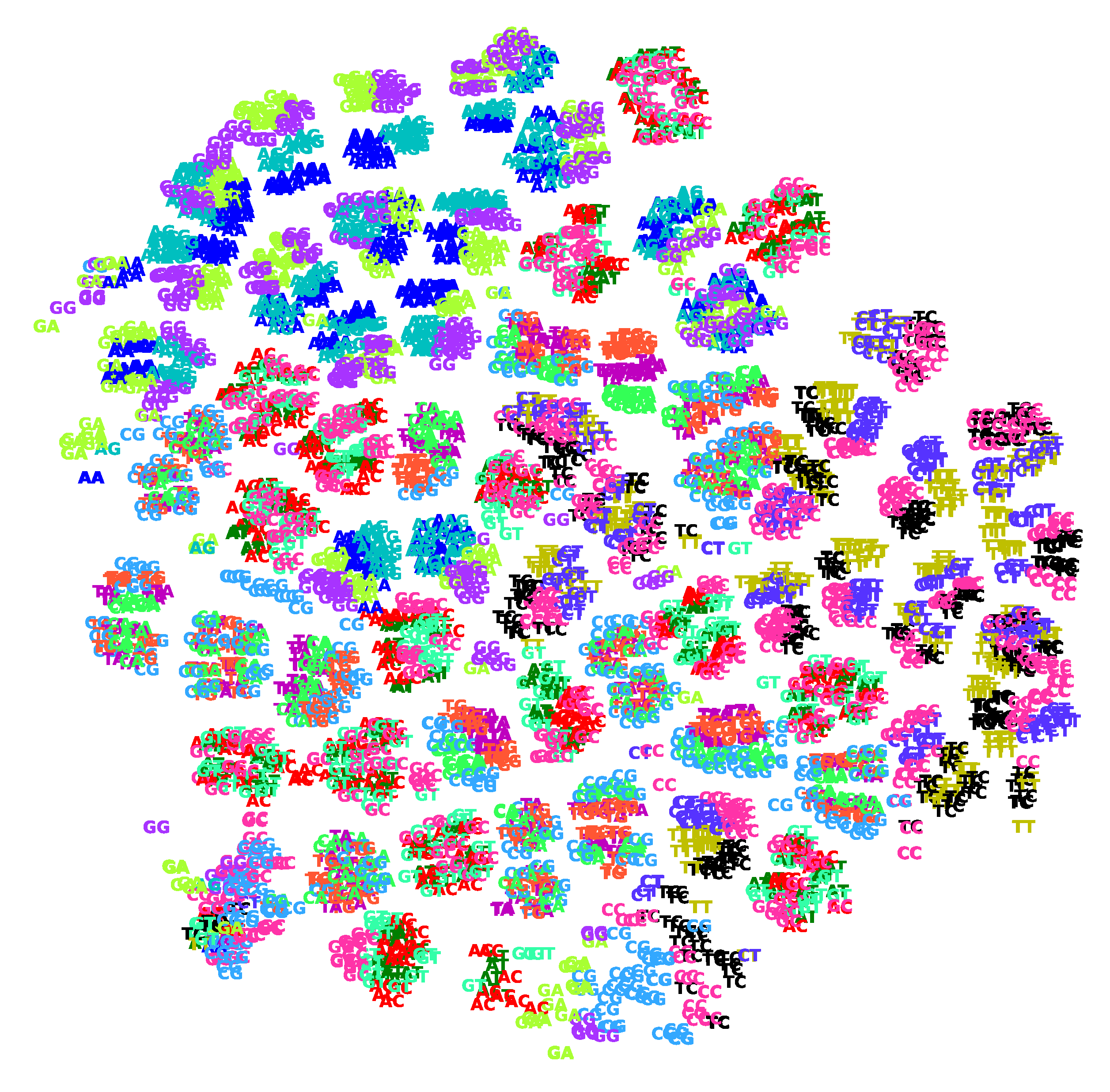}
    \vskip -0.08in
    \subcaption*{VI. 300k-th step}
  \end{minipage}
  \hfill
  % Add vertical space between the rows
  \vspace{0.1in} % Adjust this value as needed
  % 第三行的图像
  \begin{minipage}[t]{.32\linewidth}
    \centering
    \includegraphics[width=\linewidth]{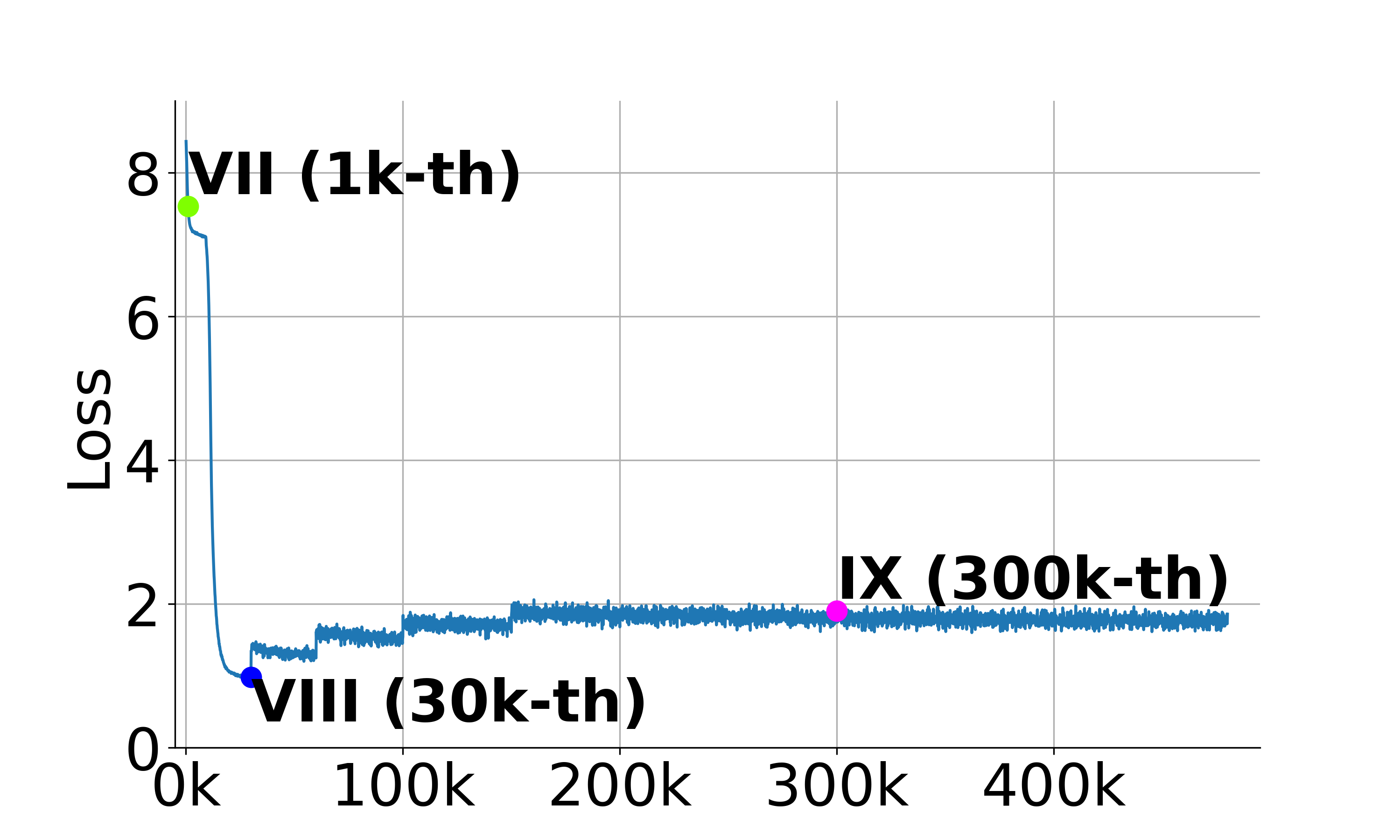}
    \vskip -0.08in
    \subcaption{The loss of Overlapping DNABERT with RandomMask (RM).}
  \end{minipage}%
  \hfill
  \begin{minipage}[t]{.22\linewidth}
    \centering
    \includegraphics[width=\linewidth]{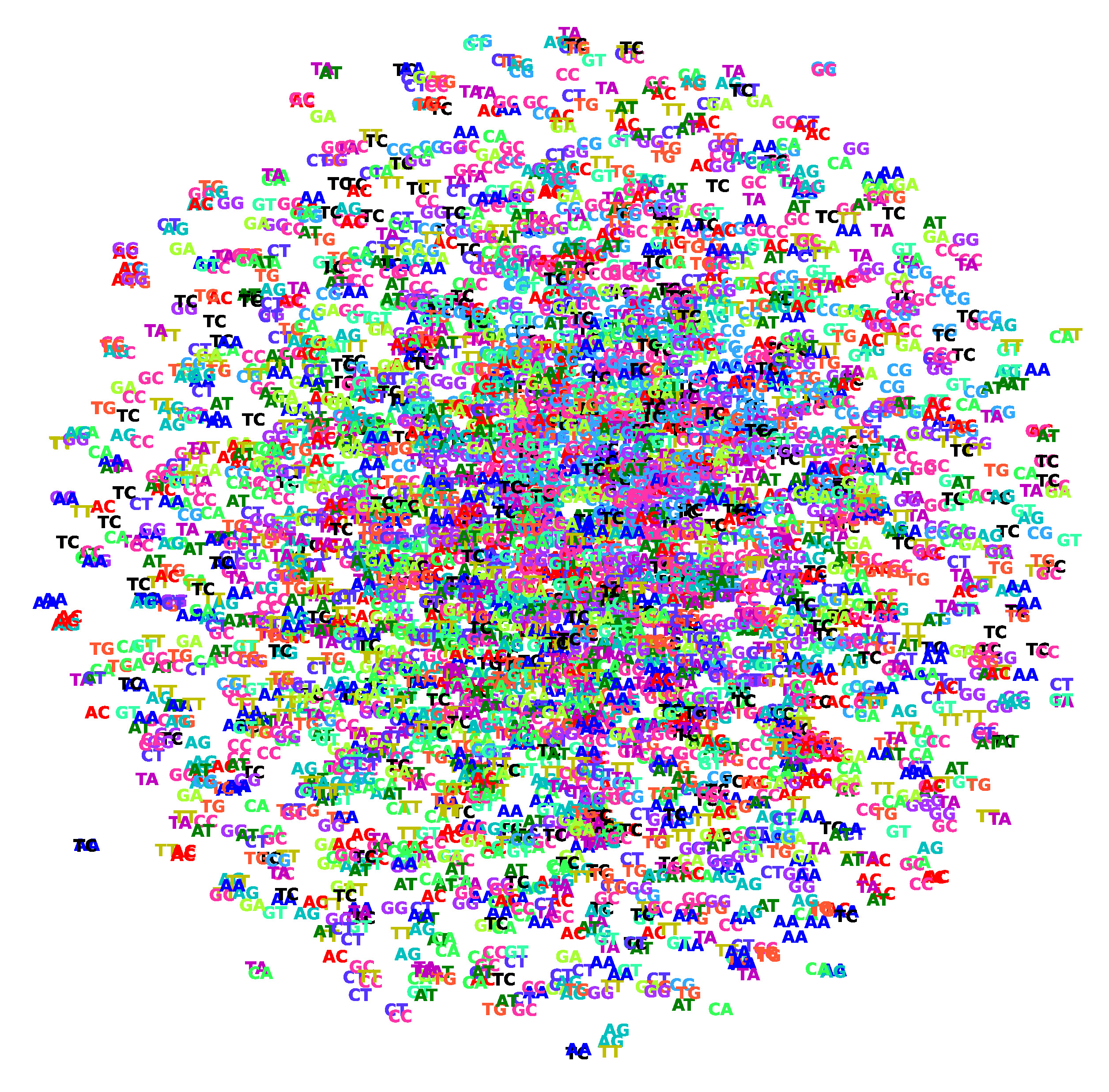}
    \vskip -0.08in
    \subcaption*{VII. 1k-th step}
  \end{minipage}
  \hfill
  \begin{minipage}[t]{.22\linewidth}
    \centering
    \includegraphics[width=\linewidth]{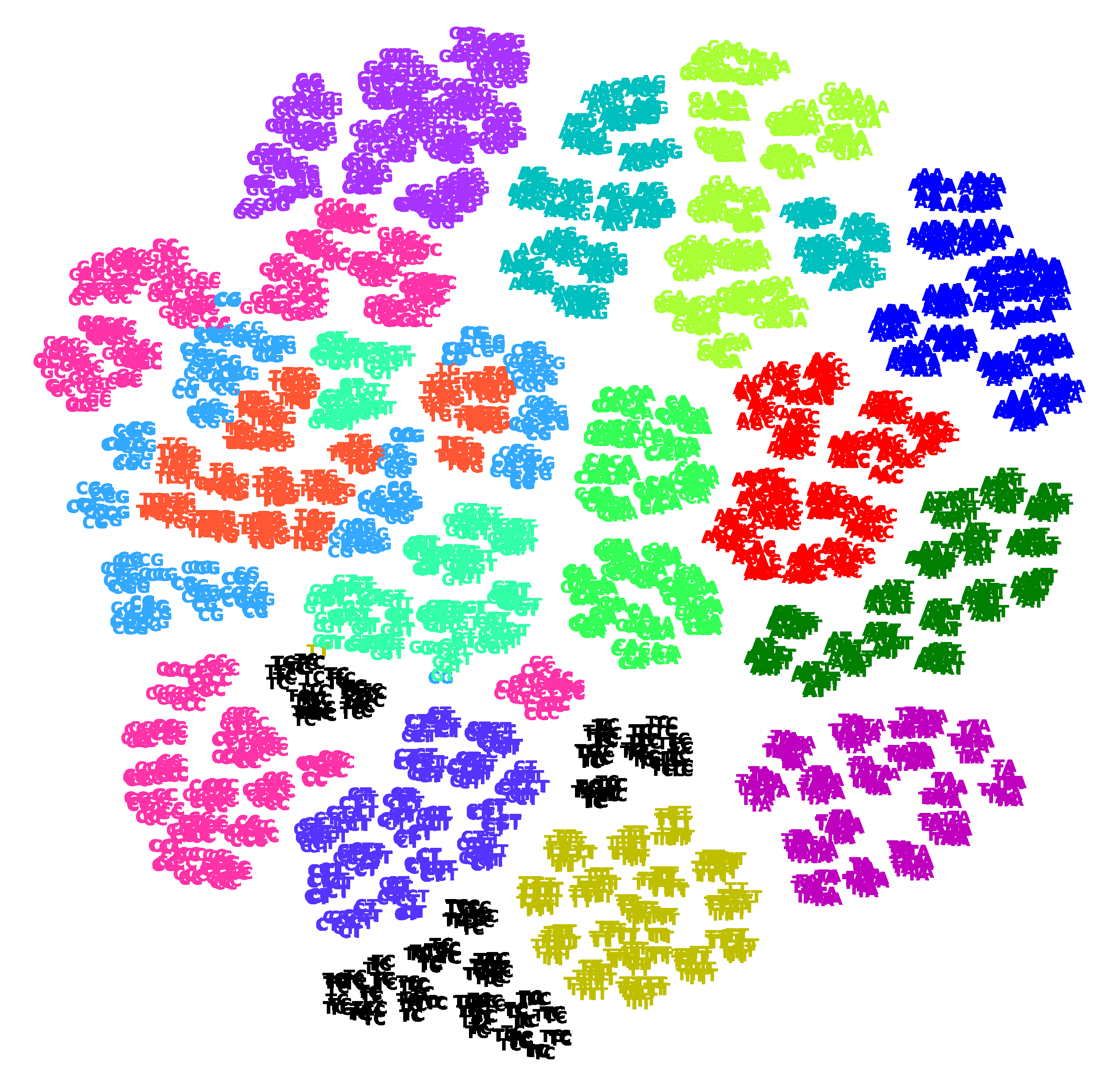}
    \vskip -0.08in
    \subcaption*{VIII. 30k-th step}
  \end{minipage}
  \hfill
  \begin{minipage}[t]{.22\linewidth}
    \centering
    \includegraphics[width=\linewidth]{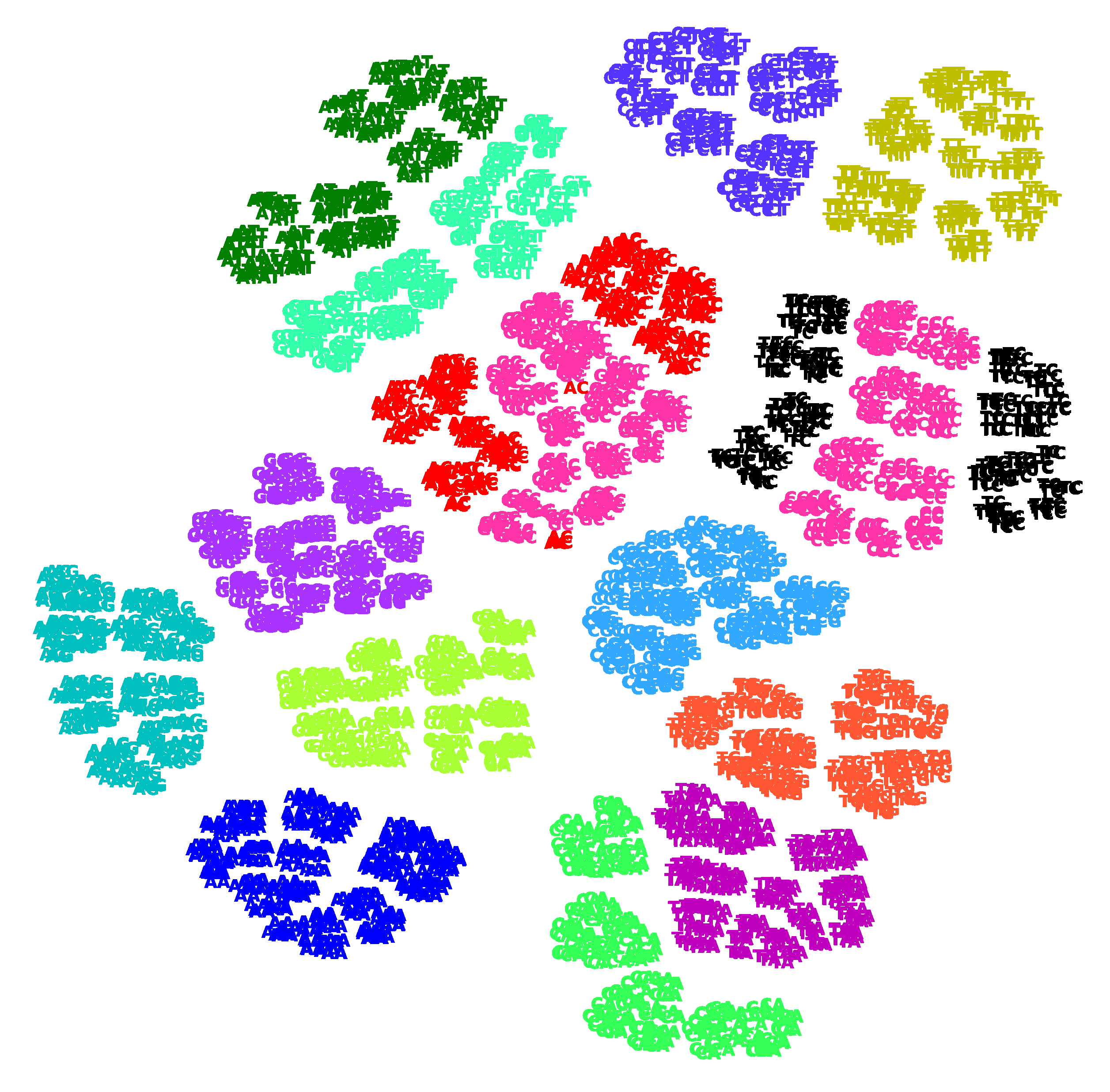}
    \vskip -0.08in
    \subcaption*{IX. 300k-th step}
  \end{minipage}
  \hfill
  \caption{The loss curves with t-SNE visualizations of the embedding spaces during the training of DNABERT with overlapping 6-mer tokenizer (the first row), DNABERT with non-overlapping 6-mer tokenizer (the second row), and Overlapping 6-mer DNABERT with RandomMask (the third row). Comparing the first and second rows, we observe that the magnitude of the loss value is inversely correlated with the level of organization observed in the embedding space. The third line shows the effect of RandomMask, which keeps the loss value high while preserving a regular arrangement of the embedding space.}
\label{figure:loss_and_cluster_comparison}
\end{figure*}
\begin{tcolorbox}[leftrule=1.5mm,top=0.8mm,bottom=0.5mm]
\textbf{Observation 2:}
\begin{itemize}
    \item 
    During the pre-training stage, overlapping tokenizer results in a more organized embedding space, rapidly reducing the loss to an exceptionally low level. Conversely, using non-overlapping tokenizer yields a less organized embedding space, with a continuous decrease in the loss.
\end{itemize}
\end{tcolorbox}
\subsubsection{Embedding Space Analysis} 
We compare the progression of embedding space and loss values between the two models. We use the t-SNE algorithm ~\cite {van2008visualizing} to visualize the embedding space and present the results in  Figure~\ref{figure:loss_and_cluster_comparison}. Comparing the two embedding spaces, we notice a notable distinction between the outcomes achieved by DNABERT when using overlapping and non-overlapping tokenizer. For overlapping tokenizer, as the loss decreases quickly, the embedding space becomes increasingly organized, resulting in a clear clustering of tokens when the loss reaches a low level. On the other hand, for non-overlapping tokenizer, the loss continuously decreases but remains relatively high, with limited organization in the embedding space.

Upon closer examination of Figure~\ref{figure:detailed_cluster}, we observe that each major cluster corresponds to the clustering of the central two nucleotides of each token, and the marginal nucleotides determine the distribution of tokens within the cluster. We refer to these two central nucleotides in each token as the ``representative elements'' of the token. These representative elements establish the crucial one-to-one correspondence between tokens and nucleotides, which is the key factor contributing to the superior performance of overlapping tokenizer.

We now give an intuitive analysis of the convergence of the two models. The rapid convergence and exceptionally low loss value of DNABERT with overlapping tokenizer demonstrate the model's proficiency in solving the MLM task. However, it also implies that the pre-training task leads to early overfitting. Nevertheless, The model's ability to recognize representative elements and utilize the highly organized embedding space allows it to efficiently narrow down the search scope and accurately identify masked tokens. Consequently, the model effortlessly accomplishes the original MLM task, as masking six tokens is essentially equivalent to masking a single nucleotide, which is a relatively simple task.
\begin{figure*}[t]
  \begin{minipage}[t]{.238\linewidth}
    \centering
    \includegraphics[width=\linewidth]{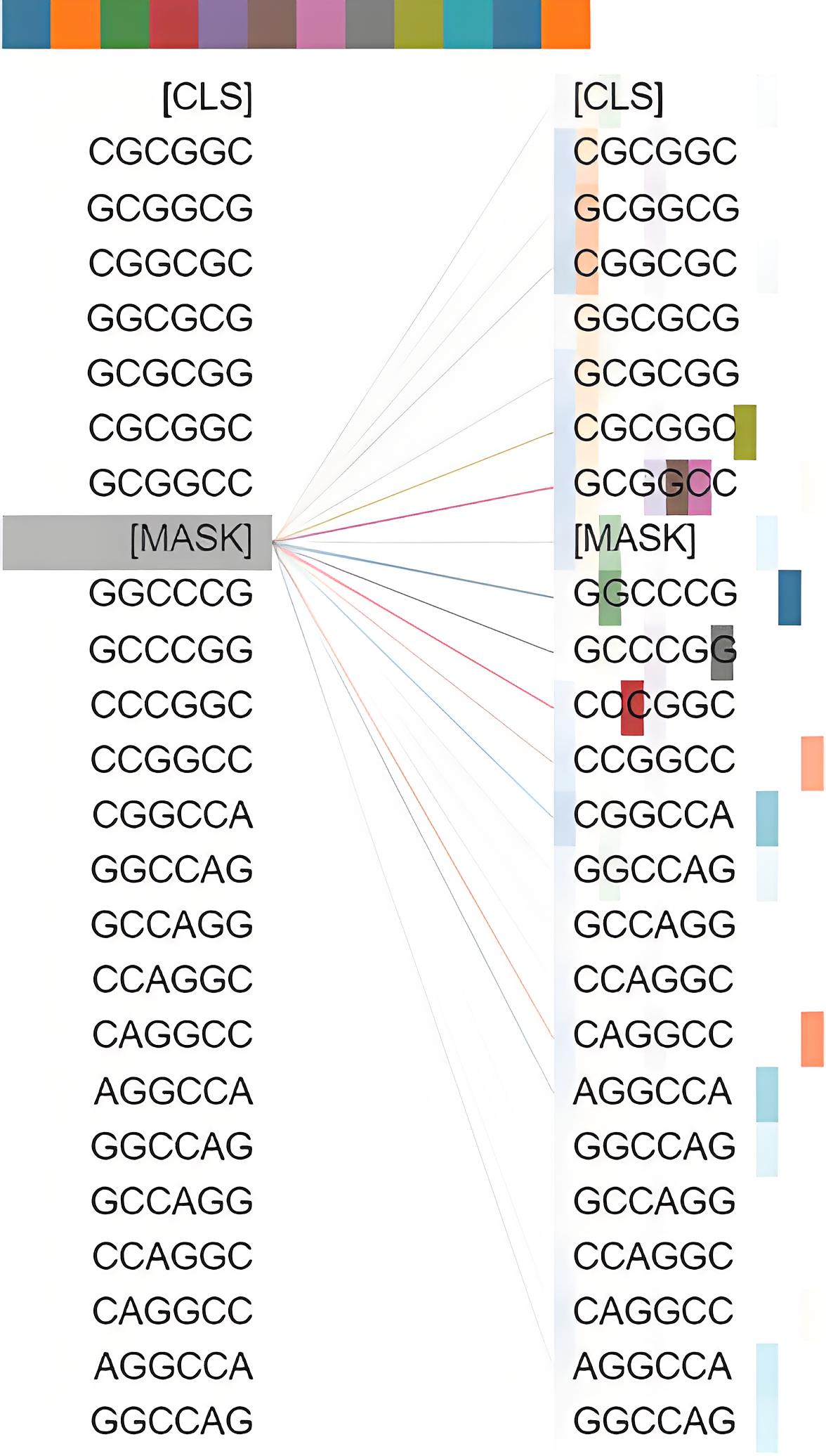}
    % \vskip -0.03in
    \subcaption{Overlapping DNABERT's last (12th) attention layer.}
  \end{minipage}%
  \hfill
  \begin{minipage}[t]{.238\linewidth}
    \centering
    \includegraphics[width=\linewidth]{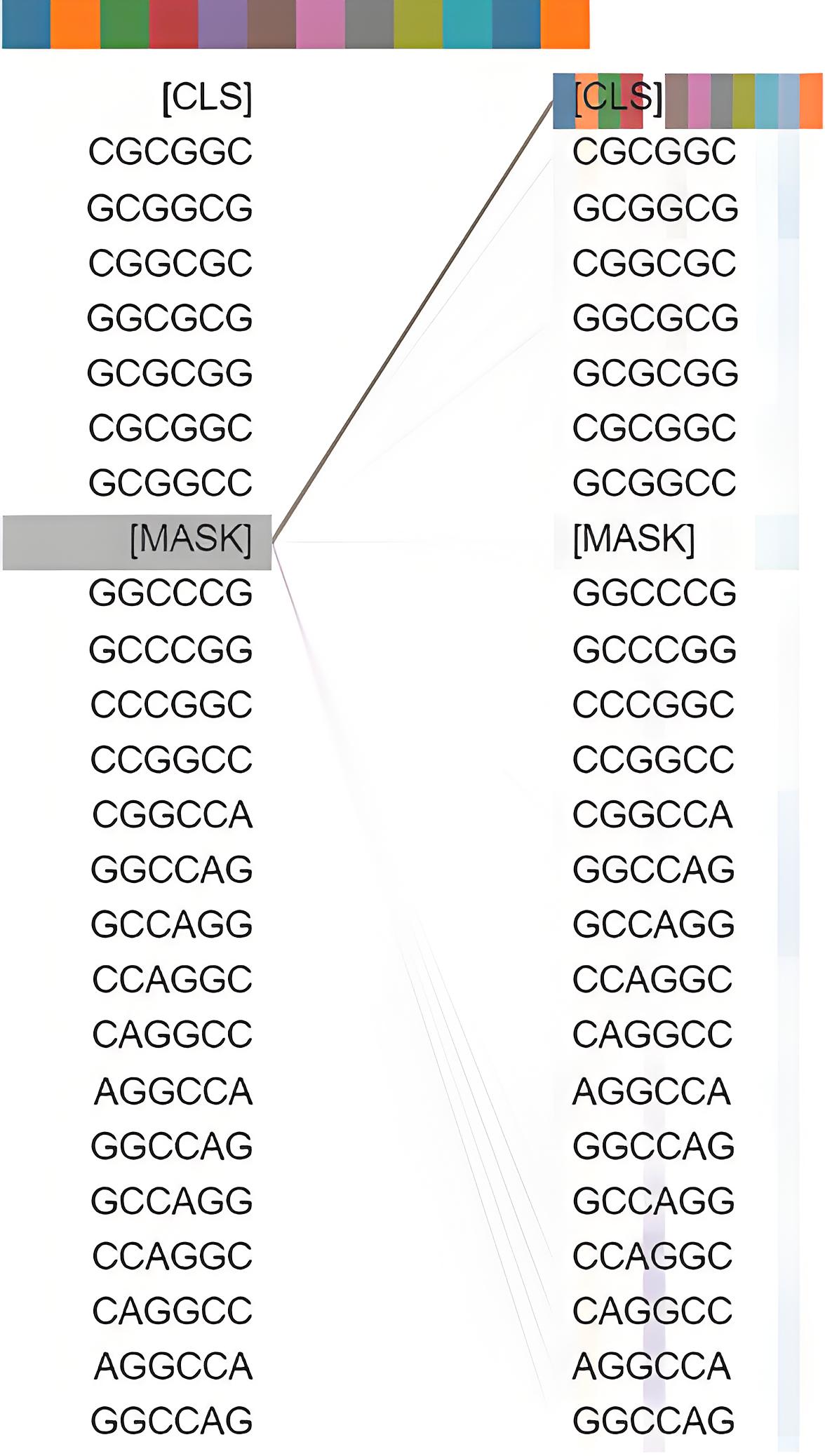}
    % \vskip -0.03in
    \subcaption{Overlapping DNABERT's intermediate (5th) attention layer.}
  \end{minipage}
  \hfill
  \begin{minipage}[t]{.24\linewidth}
    \centering
    \includegraphics[width=\linewidth]{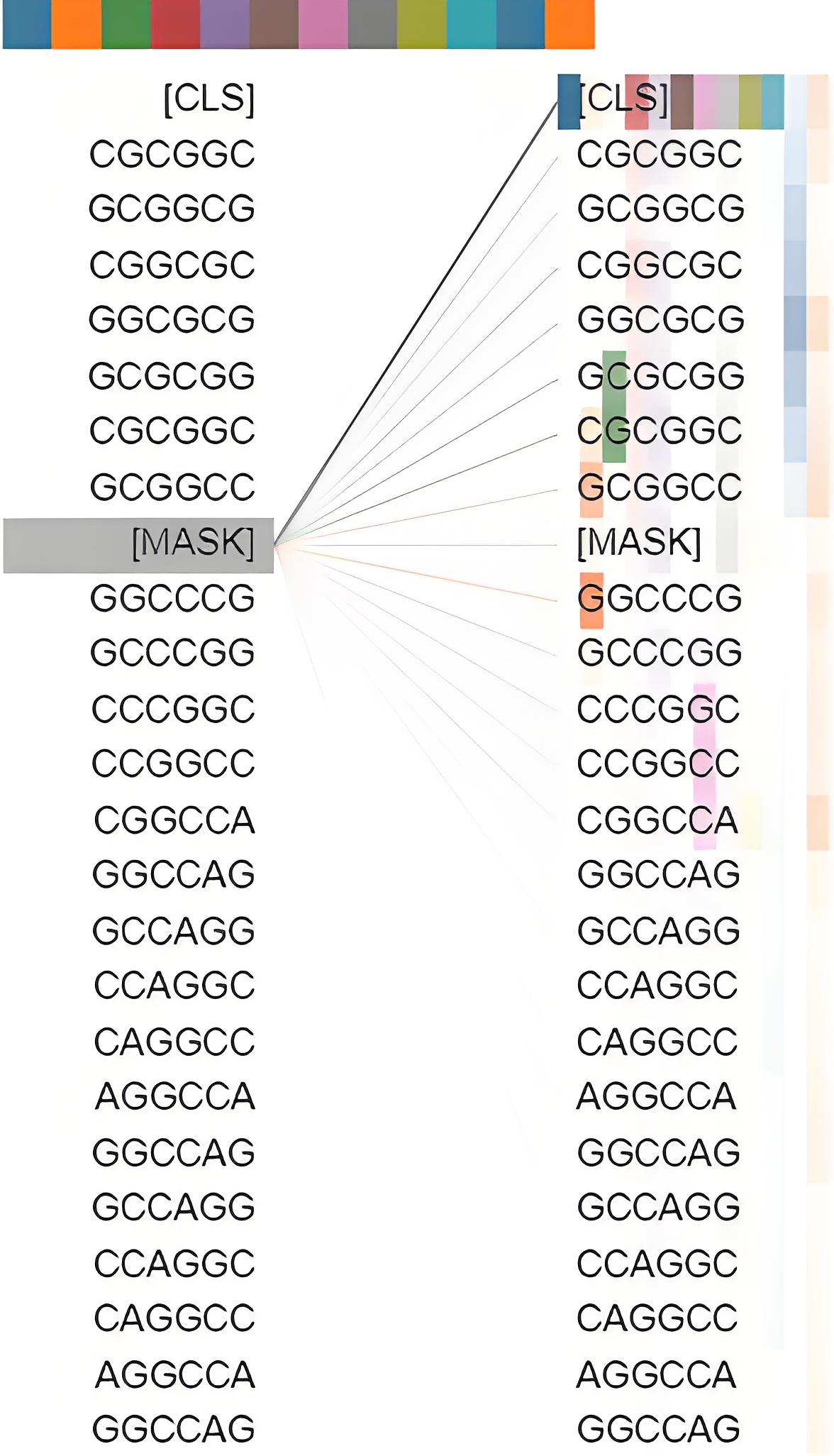}
    % \vskip -0.03in
    \subcaption{Non-overlapping DNABERT's intermediate (5th) attention layer.}
  \end{minipage}
  \hfill
  \begin{minipage}[t]{.24\linewidth}
    \centering
    \includegraphics[width=\linewidth]{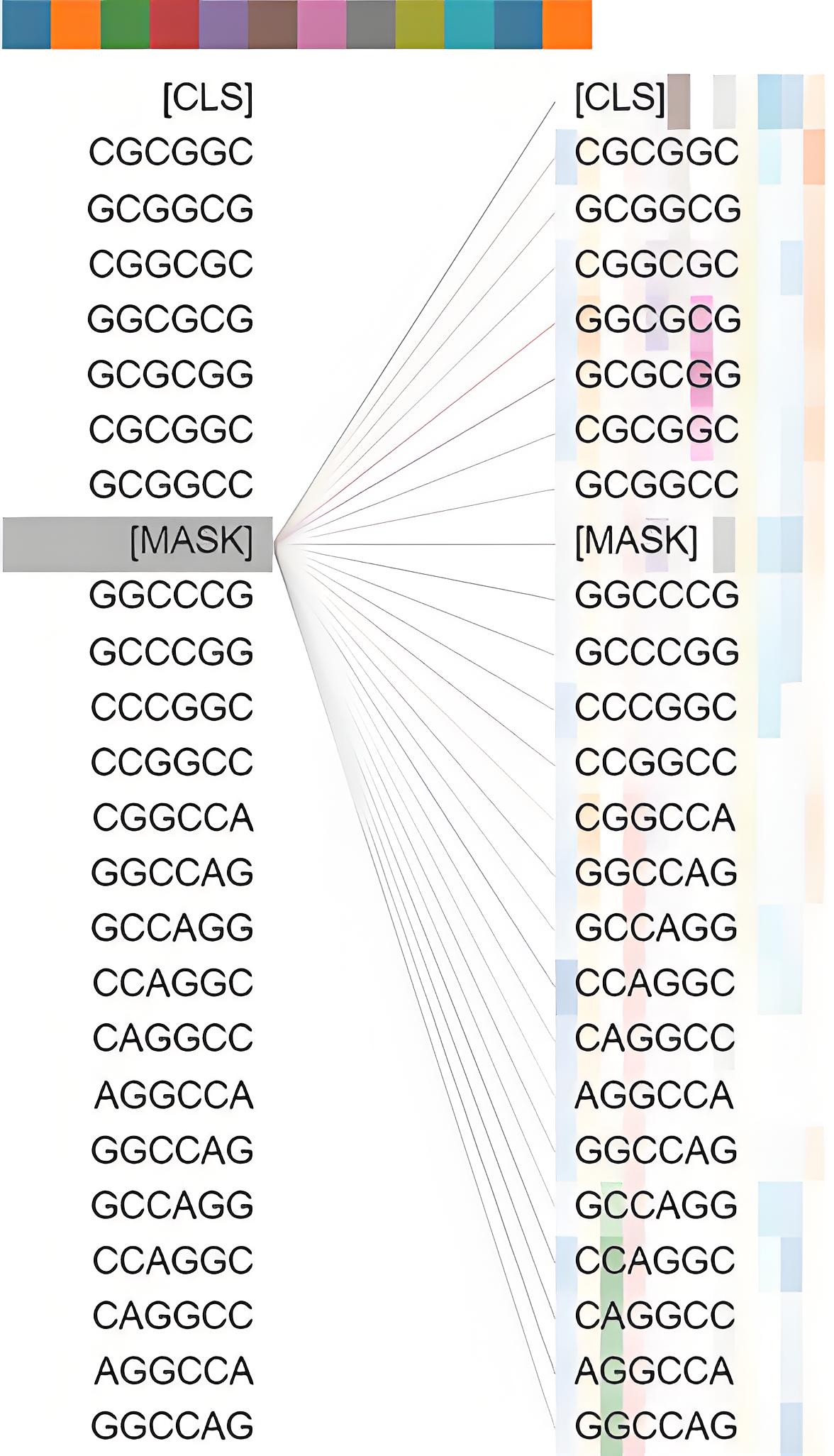}
    % \vskip -0.03in
    \subcaption{Overlapping + RM model’s intermediate (5th) attention layer.}
  \end{minipage}
  \hfill
\caption{The attention mechanism of DNABERT with overlapping 6-mer tokenizer (a, b), non-overlapping 6-mer tokenizer (c), and overlapping 6-mer DNABERT training with RandomMask (d). The 12 color blocks in the figure represent each of the 12 self-attention heads, with darker colors representing greater attention weights. From (a), it can be seen that [MASK] in the last layer of overlapping DNABERT pays diverse attention to the surrounding tokens. In (b), it can be seen that [MASK] in the middle (5th) layer of overlapping DNABERT focuses its attention on [CLS]. However, [MASK] in the middle (5th) layer of Non-overlapping DNABERT in (c) still shows diverse attentional patterns to the surrounding tokens. It suggests a potential lack of training in the middle (5th) layer of overlapping DNABERT. In (d), RandomMask is applied to solve the potential lack of training problem of overlapping DNABERT. The above diagram of the self-attention mechanism was drawn using BertViz.}
\label{figure:attention}
\end{figure*}

\subsubsection{Attention Analysis}
As previously discussed, the rapid convergence and exceptionally low loss value of DNABERT with overlapping tokenizer imply that the original MLM task is too simple for the model. This raises the possibility that the model has not been extensively trained, potentially limiting its ability to reach its full potential. In this section, we delve deeper into the analysis of the behavior of both models to validate the proposal and gain further insights.

\begin{tcolorbox}[leftrule=1.5mm,top=0.8mm,bottom=0.5mm]
\textbf{Observation 3:}
\begin{itemize}
    \item The original overlapping tokenizer has shortcomings during MLM pre-training. The intermediate layers of the trained model excessively focus on [CLS], indicating that the intermediate layers of the model are undertrained.
\end{itemize}
\end{tcolorbox}

We visualize their attention mechanism. The results are shown in Figure~\ref{figure:attention}(a) and (b). We observe that the intermediate attention mechanisms of DNABERT with overlapping tokenizer are overly concentrated on the first token, the [CLS] token, with only the final layer focusing on a few nearby tokens. On the other hand, the attention mechanism of DNABERT with non-overlapping tokenizer is more evenly and diversely distributed across the sequence.

This phenomenon suggests that the model with overlapping tokenizer effectively learns a shortcut, whereby it only relies on the final layer to memorize a limited set of mappings from nearby tokens to the output predictions. Therefore, the intermediate layers remain mostly untrained. For a model with non-overlapping tokenizer, since the nearby tokens have no explicit information about the masked token, this shortcut is not available. 

Previous work \cite{jawahar2019does, fei2021optimizing, li2023bvit} on analyzing BERT-like architectures has shown that the diversity of attentional patterns in the middle layer of BERT is key to the model's ability to model region-level information. Thus, the under-trained middle layer of overlapping models implies a lack of ability to model region-level information.

\subsection{Summary}

Since then, the previous analysis can be summarized as follows: DNA modeling needs to consider the accurate modeling of single nucleotides and the information of the whole region. Although the poor performance of the current BERT-based overlapping DNA pre-training model \cite{ji2021dnabert} has led subsequent studies such as NT \cite{dalla2023nucleotide} and DNABERT-2 \cite{zhou2023dnabert} to abandon this tokenizer approach, our analysis suggests that the overlapping tokenizer actually contributes to the modeling of single nucleotides. The underlying reason for the poor performance of the BERT-based overlapping DNA pre-training model is that the MLM pre-training approach in traditional NLP fails to adequately train the intermediate layers of the model, thus weakening its ability to model regional information.

%Additionally, we train a BERT-mini~\citep{turc2019well}  model using the overlapping tokenizer method, successfully reducing the loss to a comparable value while utilizing only one-tenth of the parameters. This further reinforces the notion that the pre-training task is too easy for the overlapping tokenizer method.

%% file: 04Method.tex
% \begin{figure*}[h]
% \centering
% \includegraphics[width=\textwidth]{figures/bs6.png}
% \caption{ The loss curves with t-SNE visualizations of the embedding spaces during the training of DNABERT with overlapping tokenizer and RandomMask.}
% \label{figure_5}
% \end{figure*}

\section{Method}
Since our method randomly expands the masking boundaries during the MLM pre-training stage, we call it RandomMask.

\textbf{Tokenizer}: %We employ 6-mer overlapping tokenizer. In the context of DNA sequence analysis, leveraging an overlapping tokenizer paradigm provides nuanced advantages for downstream tasks. Nonetheless, this methodology might precipitate an accelerated convergence during the pre-training phase. To counteract this challenge, we introduce an new pre-training strategy.
We employ 6-mer overlapping tokenizer for both pre-training and fine-tuning, as previously outlined, due to its effectiveness in capturing a comprehensive array of DNA sequence features. However, the rapid convergence characteristic of 6-mer overlapping tokenizer during the pre-training phase may lead to lack training. This, in turn, can significantly limit the model's performance potential. To address this issue, we introduce a novel pre-training strategy. %The foundational principle of this strategy is elucidated in Figure \ref{figure_5}.

% \begin{algorithm}
% \caption{DNABERT Pre-training with RandomMask}
% \begin{algorithmic}[1]
% \Procedure{AdaptiveMaskingDNABERT}{$X$}
%     \State Initialize $steps \gets [30k, 30k, 40k, 50k, 330k]$
%     \For{$i = 1$ \textbf{to} 5}
%         \State Define $masks[i] \gets [6, 8, \ldots, 6 + 2(i-1)]$
%     \EndFor

%     \For{$p = 1$ \textbf{to} 5}
%         \For{$s = 1$ \textbf{to} $steps[p]$}
%             \State $m \gets$ uniformly select from $masks[p]$
%             \State $maskPoint \gets$ select a position in $X$
            
%             \State $startMask \gets maskPoint - m/2 + 1$
%             \State $endMask \gets maskPoint + m/2$
            
%             \For{$i = startMask$ \textbf{to} $endMask$}
%                 \State Mask token $x_i$
%             \EndFor
            
%             \State Perform one training step
%         \EndFor
%     \EndFor
% \EndProcedure
% \end{algorithmic}
% \end{algorithm}
\vspace{-2mm}
\begin{algorithm}[H]
\caption{RandomMask (RM)}
\label{algorithm_1}
\begin{algorithmic}
    \Statex {\bfseries Input:} data set $X$, step $S$, probability $P$
    \Statex Initialize empty set $MaskID$ and $masks \gets [6]$
    \Statex Initialize $steps \gets [30k, 60k, 100k, 150k, 500k]$
\end{algorithmic}
\begin{algorithmic}[1]  % [1] for line numbers
    \For{$i = 0$ {\bfseries to} $3$}
        \If{$steps[i] < S \leq steps[i+1]$}
            \State $masks \gets [6, 8, \ldots, 6 + 2(i+1)]$
        \EndIf
    \EndFor
    \State $m \gets$ uniformly select from $masks$
    \For{$i = 0$ {\bfseries to} $len(X)-1$}
        \State Generate a real number $r \sim \mathcal{U}(0, 1)$
        \If{$r \leq P$}
            \State $start \gets i - m/2 + 1$
            \State $end \gets i + m/2$
            \For{$j = start$ {\bfseries to} $end$}
                \If{$0 \leq j \leq len(X)-1$}
                    \State Add $j$ to $MaskID$
                \EndIf
            \EndFor
        \EndIf
    \EndFor
    \State {\bfseries Output:} $(X, MaskID)$
\end{algorithmic}
\end{algorithm}
\textbf{Pre-training Strategy}: To mitigate the drawbacks of overlapping tokenizer during the pre-training phase,  we propose an approach that progressively expands the masking boundary centered on the masked nucleotide. This pushes the model to learn continuously. Inspired by the curriculum learning strategy in~\cite {bengio2009curriculum}, we divided the 500k pre-training steps of DNABERT with 6-mer overlapping tokenizer into five distinct phases. The length of consecutive mask tokens is randomly chosen between the minimum and maximum values. Enhance the ability of the model to capture region-level information by allowing the model to reconstruct DNA sequences of different lengths. The minimum length of consecutive masks is set to 6, and the maximum length increases by increments of 2 at each stage. %In the initial phase, adhering to DNABERT's standard configuration, we continuously masked six tokens and executed a 30k-step training. Transitioning to the second phase, with an aim to expand the masking scope while maintaining training stability, we alternated between the original six-token masking and an extended eight-token masking, uniformly choosing between them and accomplishing another 30k steps. In the third phase, the range of masking choices further broadened to six, eight, or ten tokens, completing 40k training steps. During the fourth phase, we uniformly selected masks from choices of six, eight, ten, or twelve tokens, rounding off 50k training steps. Finally, in the concluding fifth phase, our masking options expanded to include six, eight, ten, twelve, or fourteen tokens, with an extensive training period of 330k steps.
Specifically, in the training step $S$, the $MaskID$ of a DNA tokens sequence $X = ( x_1 , x_2, \ldots , x_n )$ are obtained through Algorithm~\ref{algorithm_1}, where $P$ is a pre-defined probability value, e.g., $P=2.5\%$. Then, we can get mask tokens $\{ x_i \mid i \subseteq MaskID \}$ for MLM pre-training.

%% file: 05Experiments.tex
\section{Experiments}
We train two BERT-like DNA pre-trained models, with incorporating the RandomMask (denoted as ``+ RM'') technique. DNABERT + RM is trained on the human genome \cite{gibbs2020human}. DNABERT2 (6mer) + RM is trained on multi-species genome, following the DNABERT2 pre-training datasets \cite{zhou2023dnabert}. We evaluate the models across 6 downstream tasks. All experiments follow identical settings following DNABERT \cite{ji2021dnabert} and DNABERT2 \cite{zhou2023dnabert} to ensure a fair comparison. 
\subsection{Experimental Setup}
\textbf{Architecture:}  The backbone networks of DNABERT + RM and DNABERT2 (6mer) + RM are chosen according to the configurations used in DNABERT \cite{ji2021dnabert} and DNABERT2 \cite{zhou2023dnabert}. Each of them consists of 12 Transformer Encoder layers with 768 hidden units and 12 self-attention heads. We adopt the overlapping 6-mer tokenizer method for our models. The vocabulary size is 4,101, with 4,096 tokens representing the combinations of the four nucleotides in 6-mer arrangements, and the remaining 5 tokens are reserved for special purposes. 

\textbf{Baseline:} For a comprehensive comparison, we select the following methods as baselines. DNABERT \cite{ji2021dnabert} is an early pre-training model for DNA sequences. DNABERT is pre-trained on the human genome using an overlapping 6mer tokenizer. DNABERT2 \cite{zhou2023dnabert} is the latest improved version of DNABERT, which uses genes from several species as pre-training data. DNABERT2 also introduces Byte Pair Encoding (BPE) tokenizer for the first time in DNA sequence pre-training. All these methods greatly improve the performance of the model. Also, they provide DNABERT2 (6mer) using overlapping 6mer tokenizer. The Nucleotide Transformer (NT) \cite{dalla2023nucleotide} is a large language model of DNA sequences from Instadeep and Nvidia. NT uses a non-overlapping 6mr tokenizer. NT-500M-human indicates pre-training on the human genome using a model with a parameter count of 500 million. NT-2500M-multi indicates pre-training on the genomes of multiple species using a model with a parameter count of 2500 million. These models are open-source, and all fine-tuning hyperparameters are detailed in Appendix C.%\ref{appe_hyper}. %To ensure optimal performance, the baseline performance presented in our results are the best values obtained from both the original publications and our reproduced experiments.

\textbf{Pre-training:}  %During the pre-training stage, we incorporate the RandomMask technique and mask 15\% of the tokens in the input sequence length of 512. The AdamW \cite{loshchilov2017decoupled} optimizer is employed with a learning rate of 1e-4. 
DNABERT + RM is pre-trained on the human genome \cite{gibbs2020human} for 480k steps with a batch size of 512, typically requiring around 2 days using 8 NVIDIA Tesla A100 GPUs. DNABERT2 (6mer) and DNABERT2 (6mer) + RM are trained on the multi-species dataset \citep{zhou2023dnabert} for 500k steps with a batch size of 4096, generally taking about 7 days using 8 NVIDIA Tesla A100 GPU.

\textbf{Fine-tuning:}  The models are evaluated on 6 downstream tasks, including Epigenetic Marks Prediction (EMP) \cite{pokholok2005genome, phaml2005qualitatively}, Transcription Factor Prediction on human and mouse genomes (TF-H and TF-M), Promoter Detection (PD) \cite{oubounyt2019deepromoter}, Core Promoter Detection (CPD), and Splice Site Prediction (SSP) \cite{wang2019splicefinder}. These datasets are from the Genome Understanding Evaluatio (GUE) proposed by DNABERT2 \cite{zhou2023dnabert}. Hyperparameters for fine-tuning are adapted from DNABERT2 \cite{zhou2023dnabert}, The Nucleotide Transformer \cite{dalla2023nucleotide} and HyenaDNA \cite{nguyen2023hyenadna}.%The remaining tasks are drawn from Genomics Benchmarks collected by~\cite{grevsova2023genomic}. 
These tasks (EMP, TF-M, TF-H, PD, CPD, and SSP) utilize Matthew’s correlation coefficient (MCC) as the evaluation metric. %See Appendix~\ref{appe_hyper} for more details. 

\subsection{Metric}
\label{metric}
 % {Matthews Correlation Coefficient (MCC):} 
    The Matthews Correlation Coefficient (MCC) is a metric that is widely used in classification problems to evaluate the performance of models. It is defined as:
    
    {\scriptsize \[
    MCC = \frac{TP \times TN - FP \times FN}{\sqrt{(TP + FP)(TP + FN)(TN + FP)(TN + FN)}}
    \]}
    
    % \[
    % MCC = {\scriptsize \frac{TP \times TN - FP \times FN}{\sqrt{(TP + FP)(TP + FN)(TN + FP)(TN + FN)}}}
    % \]
    where:
    
    \begin{itemize}
        \item TP = Number of True Positives
        \item TN = Number of True Negatives
        \item FP = Number of False Positives
        \item FN = Number of False Negatives
    \end{itemize}
    True Positives and True Negatives represent accurate predictions of the model, while False Positives and False Negatives denote incorrect predictions.
\begin{table*}[t]
\newcolumntype{C}[1]{>{\centering\arraybackslash}p{#1}}
    \centering
    % \caption{We present here the characteristics of relevant DNA downstream tasks, including Epigenetic Marks Prediction (EMP), Transcription Factor Prediction on the Human genome and the Mouse genome (TF-H and TF-M), Promoter Detection (PD), Core Promoter Detection (CPD), Splice Site Prediction (SSP), and Enhancer Activate Prediction (EAP). CLS and REG denote the classification and regression tasks, respectively.}
    \caption{Characteristics of relevant DNA downstream tasks, including Epigenetic Marks Prediction (EMP), Transcription Factor Prediction on the Human genome and the Mouse genome (TF-H and TF-M), Promoter Detection (PD), Core Promoter Detection (CPD), Splice Site Prediction (SSP), and Enhancer Activate Prediction (EAP). %CLS denotes classification tasks and REG denotes regression tasks. 
    The check mark (\ding{51}) indicates tasks performed at the single nucleotide and regional levels.}
        \resizebox{0.75\linewidth}{!}{
            \begin{tabular}{l C{1.5cm} C{1.5cm} C{1.5cm} C{1.5cm} C{1.5cm} C{1.5cm}}
                \toprule
                {Downstream Tasks} & {EMP} & {TF-M} & {TF-H} & {PD} & {CPD} & {SSP}\\
                \midrule
                % Task type &  CLS & CLS & CLS & CLS & CLS & CLS \\
                Species &  Yeast & Mouse & Human & Human & Human & Human \\
                Sequence length &  500 & 100 & 100 & 300 & 70 & 400 \\
                Single nucleotide & \ding{51} & \ding{51} & \ding{51} & \ding{51} & \ding{51} &  \ding{51}\\
                Regional level &  - & - & - & \ding{51} & \ding{51} & - \\
                \bottomrule
            \end{tabular}
        }
    \label{tabel_1}
\end{table*}
\subsection{List of DNA Downstream Tasks}
\label{DNA_Downstream_Tasks}
Table \ref{tabel_1} highlights the importance of nucleotide and region-level information modeling in DNA downstream tasks. Below is additional information on these tasks.

\begin{enumerate}
    \item \textbf{Epigenetic Mark Prediction (EMP)}: This task aims to determine whether the input sequence is an epigenetic mark in the yeast genome, particularly focusing on the occupancy of acetylated and methylated nucleosomes. The dataset includes various histone modifications such as H3, H4, H3K9ac, H3K14ac, H4ac, H3K4me1, H3K4me2, H3K4me3, H3K36me3, and H3K79me3. Recognizing these epigenetic marks is crucial for understanding gene expression regulation, chromatin structure, and their impact on gene function.

    \item \textbf{Transcription Factor Binding Site Prediction (TF-M and TF-H)}: This task is focused on identifying whether the input sequence is a transcription factor (TF) binding site in the mouse (TF-M) or human (TF-H) genome. Accurately identifying these binding sites is essential for revealing gene regulatory networks, understanding gene expression patterns, and exploring the molecular mechanisms of diseases.

    \item \textbf{Promoter Detection (PD)}: This task aims to determine whether the input sequence is a proximal promoter region in the human genome. Proximal promoters play a critical role in initiating transcription, making their recognition important for understanding gene regulation, identifying disease-associated genetic factors, and developing gene therapy strategies.

    \item \textbf{Core Promoter Prediction (CPD)}: Similar to proximal promoter detection, this task aims to determine whether the input sequence is a core promoter region. The core promoter is located near the transcription start site (TSS) and the start codon and is essential for transcription initiation. Recognizing core promoters is important for understanding the mechanisms of gene expression initiation and its regulation across different cell types and conditions.

    \item \textbf{Splice Site Prediction (SSP)}: This task determines whether the input sequence is a splice donor or acceptor site in the human genome. Splice sites are crucial for alternative splicing, which contributes to protein diversity and plays a significant role in understanding the impact of aberrant splicing in genetic disorders. Accurate recognition of splice sites is vital for exploring gene expression diversity, understanding disease mechanisms, and developing gene editing therapies.
\end{enumerate}

\begin{table*}[t]
	\centering
    \caption{Performance of Different Methods on Six Downstream Tasks \cite{zhou2023dnabert}, reported in the metric of MCC. RM represents RandomMask. DNABERT2 + RM (Ours) use the overlapping 6mer tokenizer. The best performance results are represented by \textbf{boldface}.
	}
	\setlength{\tabcolsep}{5mm}{
	\begin{tabular}{lccccccc}\toprule
		 Models & {EMP} & {TF-M} & {TF-H} & {PD} & {CPD} & {SSP} & {Avg.} \\
		\midrule
            {NT-500M-human} \cite{dalla2023nucleotide} & 46.47 & 61.99 & 63.95 & 90.88 & 68.55 & 84.34 & 69.36 \\

            {NT-2500M-multi} \cite{dalla2023nucleotide} & 58.06 & 67.01 & 63.32 & 91.01 & 70.33 & {89.36} & 73.18 \\

            {DNABERT} \cite{ji2021dnabert} & 51.81 & 60.40 & 64.10 & {90.48} & {70.47} & 85.44 & 70.45 \\
            {HyenaDNA} \cite{nguyen2023hyenadna} & 61.01 & 60.51 & 60.41 & 91.55 & 63.97 & 81.48 & 69.82 \\
		% \midrule
            {DNABERT2} \cite{zhou2023dnabert} & 64.47 & {68.00} & {70.11} & 91.01 & {69.37} & 84.99 & 74.66 \\
            {DNABERT2 + RM (Ours)} & \textbf{68.16} & \textbf{76.28} & \textbf{70.99} & \textbf{93.12} & \textbf{75.14} & \textbf{89.91} & \textbf{78.93} \\
		\bottomrule
	\end{tabular}}
 \label{tb:main_results}
\end{table*}
\subsection{Results}
The main results are presented in Table~\ref{tb:main_results}. Our method, RandomMask, consistently outperforms the other methods, achieving state-of-the-art performance on 6 DNA downstream tasks. The additional performance on every dataset is detailed in Appendix A.%\ref{results_details}.

For instance, in the Epigenetic Marks Prediction (EMP) task, our method DNABERT2 + RM achieved an average Matthews Correlation Coefficient (MCC) of 68.16\%, surpassing the previous best SOTA by 3.69\%. In Transcription Factor Prediction (Mouse) (TF-M), our method achieved a MCC of 76.28\%, respectively, outperforming the baseline values of 66.37\% and 63.67\%. Our approach outperformed other methods for promoter detection, and core promoter detection achieved competitive performance.

In conclusion, applying the RandomMask strategy with overlapping 6mer tokenizer significantly enhances the performance across 6 DNA downstream tasks.
% \begin{table*}[t]
% 	\centering
%     \caption{Performance of Different Methods on Six Downstream Tasks \cite{zhou2023dnabert}. RM represents RandomMask. The best performance results are represented by \textbf{boldface}.
% 	}
% 	\setlength{\tabcolsep}{5mm}{
% 	\begin{tabular}{lccccccc}\toprule
% 		 Models & {EMP} & {TF-M} & {TF-H} & {PD} & {CPD} & {SSP} & {Avg.} \\
% 		\midrule
%             {NT-500M-human} \cite{dalla2023nucleotide} & 46.47 & 61.99 & 63.95 & 90.88 & 68.55 & 84.34 &  \\

%             {NT-2500M-multi} \cite{dalla2023nucleotide} & 58.06 & 67.01 & 63.32 & 91.01 & 70.33 & {89.36} &  \\

%             {DNABERT} \cite{ji2021dnabert} & 51.81 & 60.40 & 64.10 & {90.48} & {70.47} & 85.44 &  \\
%             {HyenaDNA} \cite{nguyen2023hyenadna} & 61.01 & 60.51 & 60.41 & 91.55 & 63.97 & 81.48 & \\
% 		% \midrule
%             {DNABERT2} \cite{zhou2023dnabert} & 64.47 & {68.00} & {70.11} & 91.01 & {69.37} & 84.99 & \\
%             % \textbf{DNABERT+RM} & \underline{65.83} & 67.96 & 69.73 & \underline{92.74} & 70.89 & 87.20 & \underline{69.56} & \underline{74.84} \\
%             % \midrule
%             % \textbf{DNABERT2-6mer} \cite{zhou2023dnabert} & 48.31 & 63.67 & 63.51 & 83.78 & \underline{74.91} & 77.90 & 66.37 & 68.35 \\
%             {DNABERT2 + RM (Ours)} & \textbf{68.16} & \textbf{76.28} & \textbf{70.99} & \textbf{93.12} & \textbf{75.14} & \textbf{89.91} &  \\
% 		\bottomrule
% 	\end{tabular}}
%  \label{tb:main_results}
% \end{table*}

\begin{figure}[t]
  \begin{minipage}[t]{0.95\linewidth}
    \centering
    \includegraphics[width=\linewidth]{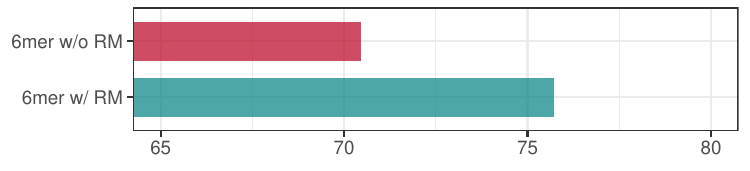}
    \vskip -0.10in
    \subcaption{DNABERT\text{  }}
  \end{minipage}
  \hfill
  \begin{minipage}[t]{0.95\linewidth}
    \centering
    \includegraphics[width=\linewidth]{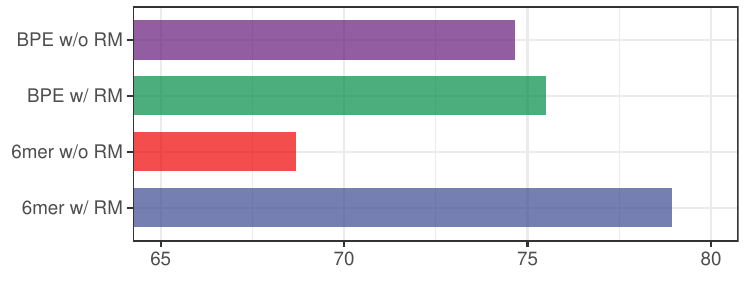}
    \vskip -0.10in
    \subcaption{DNABERT2}
  \end{minipage}
\caption{RandomMask's Ablation Study on DNABERT and DNABERT2. DNABERT and DNABERT2 (6mer) use the overlapping 6mer tokenizer. DNABERT2 (BPE) uses the BPE tokenizer. RM represents RandomMask. Our models are DNABERT + RM and DNABERT2 (6mer) + RM. It can be seen that RamdomMask is added to both DNABERT and DNABERT2 (6mer) to get performance improvement.}
\label{figure:compare}
\end{figure}

\subsection{6-mer vs BPE}

In Figure~\ref{figure:compare}, we conduct comprehensive experiments to compare our RandomMask (RM) method with DNABERT2 (BPE) and DNABERT2 (6mer)~\citep{zhou2023dnabert}. Here, DNABERT and DNABERT2 (6mer) are open-source models that use overlapping 6mer tokenizer. DNABERT2 (BPE) is an open-source model that uses BPE tokenizer.%Here we name the open-source DNABERT-2 model DNABERT2-BPE to distinguish it from the DNABERT2-6mer that we trained.
\begin{itemize}
 %%%\vspace{-2.5mm}
    \item Compare DNABERT + RM and DNABERT2. From Figure~\ref{figure:compare}, the performance of our pre-trained DNABERT + RM is slightly better than DNABERT2 (BPE).
     %%%\vspace{-0.5mm}
    \item The results in the DNABERT2 (BPE) and DNABERT2 (6mer) show that if we just replace the BPE tokenizer with the 6mer tokenizer, the model's performance will decrease. 
     %%%\vspace{-0.5mm}
    \item DNABERT2 (6mer) + RM is the model performance after using RandomMask. It can be seen that the performance of the overlapping 6mer tokenizer model has been greatly improved after using RandomMask, far exceeding DNABERT2 (BPE) and DNABERT2 (6mer).
\end{itemize}

\subsection{Model Representation Analysis}

Firstly, RandomMask obtains a clear embedding. Comparing the t-SNE plots of III, VI, and IX in Figure~\ref{figure:loss_and_cluster_comparison}, the model trained with RandomMask (IX) obtains the clearest embedding space. As stated in our analysis section, the clearer the embedding space, the more it helps improve the model's ability to model single nucleotides of DNA sequences.

Secondly, RandomMask can greatly enhance the attentional diversity of the DNABERT intermediate layer. As mentioned in our analysis section, the more diverse the model's intermediate layer attention mechanisms are represented, the better the model is at modeling regional information. By comparing Figure~\ref{figure:attention}(b), (c), and (d) of the visualization of the attentional mechanism, the model pre-trained with RandomMask (d) obtained the most diverse intermediate layer attention mechanisms. It shows that RandomMask makes the model better at modeling regional information by allowing the model to reconstruct DNA sequences of different lengths.

Thirdly, RandomMask alleviates the problem of overlapping 6mer tokenizer pre-training loss converging too fast. In Figure~\ref{figure:loss_and_cluster_comparison}, we can see that DNABERT's loss (Figure~\ref{figure:loss_and_cluster_comparison}(a)) will quickly decrease to an extremely low value. If RandomMask (Figure~\ref{figure:loss_and_cluster_comparison}(c)) is used, the loss will increase at the start of each stage, giving it enough space to decrease. We can see a decrease at each stage in the loss curve with RandomMask. RandomMask enhances the generalization of the model by increasing the difficulty of the pre-training task. %RandomMask reduces the overfitting problem caused by rapid convergence, gradually increasing the complexity during the training, which could mitigate the overfitting issue and achieve better performance by combining fast convergence and generalizability.

%% file: 07Conclusions.tex
\section{Conclusion}\

While overlapping 6-mer tokenizer offers distinct advantages in fine-tuning downstream tasks, their propensity for fast convergence can hinder comprehensive pre-training. RandomMask emerges as a potent solution, leveraging adaptive masking to push models to learn more effectively and deeply. RandomMask ensures that models can handle DNA sequences' nuances and broad patterns (nucleotide and region-level information) by continuously increasing task difficulty and expanding mask boundaries. Using RandomMask during BERT-like DNA pre-training improves the performance of the model. In particular, the performance improvement of RandomMask is more obvious when overlapping 6-mer tokenizer is used.

%% file: Appendix.tex
\newpage
\section*{Appendix}
\label{Appendix}
\begin{table*}[t]
    \centering
    \caption{Performance of Different Models on Six Benchmark Downstream Tasks. The $\uparrow$ and $\downarrow$ represent the performance improvement and degradation due to RandomMask (RM), respectively. Performance metrics are reported as MCC. The best performance results are represented by \textbf{boldface}, and the second best performance results are \underline{underlined}.}
    \resizebox{0.99\textwidth}{!}{
    \begin{tabular}{cccccccccccc}
        \toprule
        \multirow{2}{*}{{Models}} & \multicolumn{11}{c}{Epigenetic Marks Prediction (EMP)} \\
        \cmidrule(lr){2-12}
         & H3 & H3K14ac & H3K36me3 & H3K4me1 & H3K4me2 & H3K4me3 & H3K79me3 & H3K9ac & H4 & H4ac & Avg.\\
        \midrule
        {NT-500M-human [13]} & 72.60&39.11&44.25&35.47&27.59&23.49&59.14&51.39&77.07&34.54&46.47 \\
        {NT-2500M-multi [13]} & {78.77} & {56.20} & {61.99} & \textbf{55.30} & {36.49} & {40.34} & 64.70 & {56.01} & {81.67} & 49.13 & 58.06\\
        HyenaDNA [15] & 77.57 & 61.80 & 59.71 & 49.82 & {44.86} & 58.17 & 65.74& 63.37 & 74.53 & 54.50 & 61.01  \\
        DNABERT2 (BPE) [16] & \underline{81.10} & \underline{67.69} & \underline{67.57} & \underline{54.61} & 29.59 & \underline{61.81} & 72.57 & 61.92 & \textbf{82.10} & \underline{65.69} & 64.47 \\
        \hline
        DNABERT [11]& 75.82 & 48.07 & 51.52 & 43.92 & 31.01 & 37.13 & 58.98 & 52.07 & 77.85 & 41.74 & 51.81 \\
        {DNABERT+RM} & {77.62} \scriptsize($\uparrow$1.8) & {65.07} \scriptsize($\uparrow$17.0) & {63.68} \scriptsize($\uparrow$12.16) & {54.47} \scriptsize($\uparrow$10.55)& \underline{53.88} \scriptsize($\uparrow$22.87) & \textbf{62.19} \scriptsize($\uparrow$25.06) & \underline{72.67} \scriptsize($\uparrow$13.69)& \underline{65.02} \scriptsize($\uparrow$12.95) & {79.44} \scriptsize($\uparrow$1.59) & {64.22} \scriptsize($\uparrow$22.48) & \underline{65.83} \scriptsize($\uparrow$14.02)\\
        \hline
        DNABERT2 (6mer) [16] & 74.62&	42.71&	47.26&	39.66&	25.33&	27.43 & 61.03 &49.35&	78.61&	37.14 & 48.31 \\
        {DNABERT2 (6mer)+RM} & \textbf{81.87} \scriptsize($\uparrow$7.25) & \textbf{68.79} \scriptsize($\uparrow$26.08) & \textbf{68.60} \scriptsize($\uparrow$21.34) & {54.15} \scriptsize($\uparrow$14.49) & \textbf{54.09} \scriptsize($\uparrow$28.76) & {61.12} \scriptsize($\uparrow$33.69) & \textbf{75.30} \scriptsize($\uparrow$14.27)&\textbf{68.70} \scriptsize($\uparrow$19.35) & \underline{81.81} \scriptsize($\uparrow$3.20) & \textbf{67.17} \scriptsize($\uparrow$30.03) & \textbf{68.16}\scriptsize($\uparrow$19.85) \\
        \bottomrule
    \end{tabular}
    }
    \vspace{+3.mm}
    
    % \resizebox{0.80\textwidth}{!}{
    % \begin{tabular}{ccccccc}

    %     \toprule
    %     \multirow{2}{*}{{Models}} & \multicolumn{4}{c}{Epigenetic Marks Prediction (EMP)} \\
    %     \cmidrule(lr){2-5}
    %     & H3K9ac & H4 & H4ac & avg. & Enhancer & Splice Site \\
    %     \midrule
        
    %     {NT-500M-human [13]} &51.39&77.07&34.54&46.47&64.67&84.34 \\
    %     {NT-2500M-multi [13]} & {56.01} & {81.67} & 49.13 & 58.06 & 67.31 & \textbf{89.36}\\
    %     HyenaDNA [15] & 63.37 & 74.53 & 54.50 & 61.01 & 63.34 & 81.48\\
    %     DNABERT2 (BPE) [16] & 61.92 & \textbf{82.10} & \underline{65.69} & 64.47 & 67.79 & 84.99\\
    %     \hline
    %     DNABERT [11]& 52.07 & 77.85 & 41.74 & 51.81 & 68.43 & 85.44  \\
    %     {DNABERT+RM} & \underline{65.02} \scriptsize($\uparrow$12.95) & {79.44} \scriptsize($\uparrow$1.59) & {64.22} \scriptsize($\uparrow$22.48) & \underline{65.83} \scriptsize($\uparrow$14.02) & \underline{69.56} \scriptsize($\uparrow$1.13) & {87.20} \scriptsize($\uparrow$1.76)\\
    %     \hline
    %     DNABERT2 (6mer) [16] &49.35&	78.61&	37.14 & 48.31 & 66.37 & 77.90\\
    %     % \textbf{DNABERT2 (6mer)+RM} & \textbf{68.70} & \underline{81.81} & \textbf{67.17} & \textbf{68.16} & \textbf{70.41} & \underline{88.91}\\
    %     {DNABERT2 (6mer)+RM} & \textbf{68.70} \scriptsize($\uparrow$19.35) & \underline{81.81} \scriptsize($\uparrow$3.20) & \textbf{67.17} \scriptsize($\uparrow$30.03) & \textbf{68.16} \scriptsize($\uparrow$19.85) & \textbf{70.41} \scriptsize($\uparrow$4.04) & \underline{89.91} \scriptsize($\uparrow$12.01) \\
    %     \bottomrule
    % \end{tabular}
    % }

    % \vspace{+3.mm}
    
    \resizebox{0.75\textwidth}{!}{
    \begin{tabular}{
        >{\centering\arraybackslash}p{2.9cm}
        *{3}{>{\centering\arraybackslash}p{1.6cm}}
        *{3}{>{\centering\arraybackslash}p{1.6cm}}
        >{\centering\arraybackslash}p{1.6cm}
    }
        \toprule
        \multirow{2}{*}{{Models}} & \multicolumn{3}{c}{Core Promoter Detection} & \multicolumn{3}{c}{Promoter Detection} & \multirow{2}{*}{Splice Site} \\
        \cmidrule(lr){2-4}
        \cmidrule(lr){5-7}
        & notata & tata & all & notata & tata & all \\
        \midrule
        {NT-500M-human [13]} &68.71&73.90&68.55& 93.37&80.49 & 90.88&84.34\\
        {NT-2500M-multi [13]} &\underline{71.58} & 72.97 & {70.33} &\text{94.00} & {79.43}&91.01& \underline{89.36}\\
        HyenaDNA [15] & 63.77 & 64.16 & 63.97 & 85.14 & 53.19 & 91.55 & 81.48\\
        DNABERT2 (BPE) [16] & 68.04 & 74.17 & 69.37& \text{94.00} &79.34&91.01& 84.99\\
        \hline
        DNABERT [11]& \textbf{71.88} & 76.06 & {70.47}& 93.05 & 61.56 & 90.48& 85.44\\
        % \textbf{DNABERT+RM} & {67.13} & {72.55} & \textbf{71.64} & \textbf{60.14} & {77.20} & {69.73} & {93.40} & \underline{84.03} & \textbf{92.74} \\
        {DNABERT+RM} & 71.50 \scriptsize($\downarrow$0.38) & \underline{76.65} \scriptsize($\uparrow$0.59) & {70.89} \scriptsize($\uparrow$0.42) & {93.40} \scriptsize($\uparrow$0.35) & \textbf{84.03} \scriptsize($\uparrow$22.47) & \underline{92.74} \scriptsize($\uparrow$2.26) & {87.20} \scriptsize($\uparrow$1.76)\\
        \hline
        DNABERT2 (6mer) [16] & {69.23} & {74.91} & \underline{74.91}& {92.65} & 57.75 & 83.78& 77.90\\
        % \textbf{DNABERT2 (6mer)+RM}  & \underline{70.78} & \underline{72.82} & \underline{70.69} & {57.80} & \textbf{82.84} & \textbf{70.99} & \textbf{94.42} & \textbf{85.37} & \underline{92.68} \\
        {DNABERT2 (6mer)+RM} & 70.27 \scriptsize($\uparrow$1.04) & \textbf{78.51} \scriptsize($\uparrow$3.60) & \textbf{75.14} \scriptsize($\uparrow$0.23)  & \underline{93.55} \scriptsize($\uparrow$0.90) & \underline{83.03} \scriptsize($\uparrow$25.28) & \textbf{93.12} \scriptsize($\uparrow$2.34)& \textbf{89.91} \scriptsize($\uparrow$12.01)\\
        \bottomrule
    \end{tabular}
    }
    \vspace{+3.mm}

    \resizebox{0.75\textwidth}{!}{
    \begin{tabular}{
        >{\centering\arraybackslash}p{2.9cm}
        *{6}{>{\centering\arraybackslash}p{1.6cm}}
    }
        \toprule
        \multirow{2}{*}{{Models}} & \multicolumn{6}{c}{Transcription Factor Prediction (Human)}\\
        \cmidrule(lr){2-7}
        & 0 & 1 & 2 & 3 & 4 & Avg.\\
        \midrule
        {NT-500M-human [13]} & 66.95&67.29&62.20&47.29&76.03&63.95 \\
        {NT-2500M-multi [13]} &  66.64 & 70.28 & 58.72 & 51.65 & 69.34 &63.32\\
        HyenaDNA [15] & 60.96 & 56.68 & 60.66 & 51.01 & 72.73 & 60.41 \\
        DNABERT2 (BPE) [16] & \textbf{71.99} &\textbf{76.06}&66.52&\underline{58.54}&\textbf{77.43}&\textbf{70.11}\\
        \hline
        DNABERT [11]& 67.06 & 69.83 & 61.78 & 47.08 & 74.77 & 64.10\\
        % \textbf{DNABERT+RM} & {67.13} & {72.55} & \textbf{71.64} & \textbf{60.14} & {77.20} & {69.73} & {93.40} & \underline{84.03} & \textbf{92.74} \\
        {DNABERT+RM} & {67.13} \scriptsize($\uparrow$0.07) & {72.55} \scriptsize($\uparrow$2.72) & \textbf{71.64} \scriptsize($\uparrow$9.86) & \textbf{60.14} \scriptsize($\uparrow$13.06) & \underline{77.20} \scriptsize($\uparrow$2.43) & \underline{69.73} \scriptsize($\uparrow$5.63) \\
        \hline
        DNABERT2 (6mer) [16] & {67.99}	&67.06	&59.45	&50.24&	{72.80} & 63.51\\
        % \textbf{DNABERT2 (6mer)+RM}  & \underline{70.78} & \underline{72.82} & \underline{70.69} & {57.80} & \textbf{82.84} & \textbf{70.99} & \textbf{94.42} & \textbf{85.37} & \underline{92.68} \\
        {DNABERT2 (6mer)+RM} & \underline{70.78} \scriptsize($\uparrow$2.79) & \underline{72.81} \scriptsize($\uparrow$5.75) & \underline{67.18} \scriptsize($\uparrow$7.73) & {52.91} \scriptsize($\uparrow$2.67) & {75.26} \scriptsize($\uparrow$2.46) & {66.17} \scriptsize($\uparrow$2.66) \\
        \bottomrule
    \end{tabular}
    }
    \vspace{+3.mm}

        \resizebox{0.75\textwidth}{!}{
    \begin{tabular}{
        >{\centering\arraybackslash}p{2.9cm}
        *{6}{>{\centering\arraybackslash}p{1.6cm}}
    }
        \toprule
        \multirow{2}{*}{{Models}} & \multicolumn{6}{c}{Transcription Factor Prediction (Mouse)} \\
        \cmidrule(lr){2-7}
        & 0 & 1 & 2 & 3 & 4 & Avg.\\
        \midrule
        {NT-500M-human [13]} & 50.54&77.73&{78.05}&61.01&42.64&61.99\\
        {NT-2500M-multi [13]} & \underline{63.31} & 83.76 & 71.52 & {69.44} & 47.07 &67.01\\
        HyenaDNA [15] & 47.55 & 79.85 & 74.58 & 58.77 & 41.81 & 60.51 \\
        DNABERT2 (BPE) [16] & 56.76 & \underline{84.77} & \underline{79.32} & 66.47 & \underline{52.66} & \underline{68.00}\\
        \hline
        DNABERT [11]& 46.27 & 78.84 & 74.41 & 59.04 & 43.45 & 60.40 \\
        % \textbf{DNABERT+RM} & {55.61} & {82.72} & 77.61 & \underline{74.06} & {49.81} & {67.96} & 71.50 & \underline{76.65} & \underline{70.89} \\
        {DNABERT+RM} & {55.61} \scriptsize($\uparrow$9.34) & {82.72} \scriptsize($\uparrow$3.88) & 77.61 \scriptsize($\uparrow$3.20) & \underline{74.06} \scriptsize($\uparrow$15.02) & {49.81} \scriptsize($\uparrow$6.36) & {67.96} \scriptsize($\uparrow$7.56) \\
        \hline
        DNABERT2 (6mer) [16] &{48.96} &	81.69&	81.71	&63.17	&42.83 & 63.67 \\
        % \textbf{DNABERT2 (6mer)+RM}  & \textbf{70.00} & \textbf{85.77} & \textbf{85.99} & \textbf{85.80} & \textbf{53.85} & \textbf{76.28} & 70.27 & \textbf{78.51} & {75.14} \\
        {DNABERT2 (6mer)+RM} & \textbf{70.00} \scriptsize($\uparrow$21.04) & \textbf{85.77} \scriptsize($\uparrow$4.08) & \textbf{85.99} \scriptsize($\uparrow$4.28) & \textbf{85.80} \scriptsize($\uparrow$22.63) & \textbf{53.85} \scriptsize($\uparrow$11.02) & \textbf{76.28} \scriptsize($\uparrow$12.61) \\
        
        \bottomrule
    \end{tabular}
    }
    \label{table:detail_main_result}
\end{table*}
\begin{table*}[t]
    \centering
    \caption{Expanded comparison of different tokenizer strategies for DNABERT and NT across 6 downstream tasks [16]. Strategies include Non-overlapping, Same-length, and Overlapping tokenizer. Performance metrics are reported as MCC. The best performance results are represented by \textbf{boldface}, and the second best performance results are \underline{underlined}.}
    \label{table_expanded_comparison}
    \resizebox{0.70\textwidth}{!}{
        \begin{tabular}{@{}llcccccccl@{}}
            \toprule
            {Model} & {tokenizer} & {EMP} & {TF-M} & {TF-H} & {PD} & {CPD} & {SSP}  & Avg. \\
            \midrule
            \multirow{3}{*}{NT [13]} & Non-overlapping & \underline{45.37} & 39.81 & 55.25 & \underline{88.43} & 62.56 & 80.39 & 61.97 \\
                                & Same-length & 44.88 & \underline{47.59} & \underline{60.57} & 86.96 & \underline{63.98} & \underline{80.96} & \underline{64.16} \\
                                & Overlapping & \textbf{46.47} & \textbf{61.99} & \textbf{63.95} & \textbf{90.88} & \textbf{68.55} & \textbf{84.34} & \textbf{69.36} \\
            \midrule
            \multirow{3}{*}{DNABERT [11]} & Non-overlapping & \underline{43.65} & 34.87 & \underline{54.50} & \underline{87.62} & \underline{65.82} & 79.91 & \underline{61.06} \\
                                     & Same-length & 42.98 & \underline{38.60} & 53.27 & 85.33 & 64.09 & \underline{80.76} & 60.84 \\
                                     & Overlapping & \textbf{51.81} & \textbf{59.60} & \textbf{63.55} & \textbf{90.48} & \textbf{70.47} & \textbf{85.44} & \textbf{70.23} \\
            \bottomrule
        \end{tabular}
    }
\end{table*}

\subsection{Additional Results}
\label{results_details}
Table \ref{table:detail_main_result} shows the results for each dataset on the 7 downstream tasks.

\begin{figure}[t]
\includegraphics[width=0.45\textwidth]{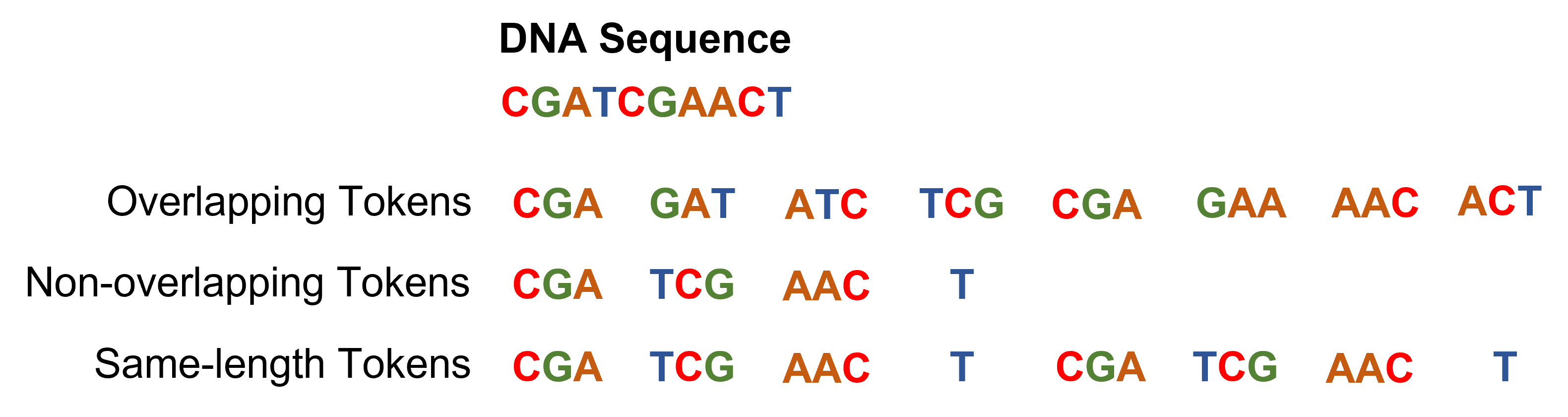}
% \vspace{-1mm}
\caption{Examples for 3-mer DNA token with Overlapping, Non-overlapping, and Same-length strategies. Same-length and overlapping are the same token sequence length.}
\label{figure:same_length_4}
\end{figure}

\subsection{Sensitivity Analysis on Sequence Length}
In Table~\ref{table_expanded_comparison}, we investigate the effect of sequence length. The row labeled ``Same-length" shows the effect of creating a sequence with the same length as the ``Overlapping" sequence by repeating the tokens from the "Non-overlapping" sequence K-1 times.

Examples shown in Figure \ref{figure:same_length_4}, if the ``Non-overlapping" sequence is token by 3-mer ``$token_1$ $token_2$," the ``Same-length" sequence would be ``$token_1$ $token_2$ $token_1$ $token_2$." In other words, the "Non-overlapping" sequence ``$token_1$ $token_2$" is repeated 2 times to match the length of the ``Overlapping" sequence.

This method allows us to compare the performance of non-overlapping and overlapping sequences of the same length. The judgments of the comparison are displayed as follows:
\begin{itemize}
    \item An interesting phenomenon. In NT that uses non-overlapping 6-mer for pre-training, stretching the sequence length will indeed produce obvious gains in TF-M, TF-H, and CPD. Combined with Table 1 in the paper, the common feature of these three tasks is that the DNA sequence length is short. The DNA sequence lengths of EMP, PD, and SSP are 500, 300, 400, and 250 nucleotides, respectively. However, the DNA sequence lengths of TF-M, TF-H, and CPD are 100, 100, and 70 nucleotides, respectively, and these are shorter than others.
    \item But in general, using the overlapping tokenizer to obtain more diverse tokens achieves better performance than simply lengthening the sequence length (Same-length) of both the overlapping pre-training (DNABERT) model and the non-overlapping pre-training model (NT).
\end{itemize}

\subsection{Hyperparameters}
\label{appe_hyper}
Table \ref{hyper1} summarizes the default hyperparameter settings for various configurations of the DNABERT models.%, including DNABERT+RM, DNABERT2 (BPE), DNABERT2 (6mer), and DNABERT2 (6mer)+RM.
\begin{table}[h]
    \centering
    \captionsetup{font=small}
    \caption{Default hyperparameter settings for DNABERT and DNABERT+RM, DNABERT2 (BPE), DNABERT2 (6mer) and DNABERT2 (6mer)+RM in downsteam tasks.}
    %%\vspace{-2mm}
    \resizebox{0.8\linewidth}{!}{
        \begin{tabular}{lccccc}
            \toprule
             Downstream Task & EMP & TF & CPD & PD & SSP \\
            \midrule
            Optimizer & \multicolumn{5}{c}{AdamW} \\
            Optimizer momentum & \multicolumn{5}{c}{$\beta_1$, $\beta_2$ = 0.9, 0.999} \\
            Batch size & 32 & 32 & 32 & 32 & 32 \\
            Training epoch & 100 & 10 & 10 & 5 & 10 \\
            \midrule
            Learning rate & \multicolumn{5}{c}{3e-5}\\
            Weight decay & \multicolumn{5}{c}{0}\\
            \bottomrule
        \end{tabular}
    }
    \label{hyper1}
\end{table}

Table \ref{hyper2} presents the default hyperparameter settings for the Nucleotide Transformer model across various downstream tasks. %The settings for this model slightly differ from those of DNABERT, particularly in the learning rate parameter, to better accommodate the unique architecture and training dynamics of the Nucleotide Transformer.
\begin{table}[h]
    \centering
    \captionsetup{font=small}
    \caption{Default hyperparameter settings for the Nucleotide Transformer in downsteam tasks.}
    %\vspace{-2mm}
    \resizebox{0.8\linewidth}{!}{
        \begin{tabular}{lccccc}
            \toprule
             & EMP & TF & CPD & PD & SSP \\
            \midrule
            Optimizer & \multicolumn{5}{c}{AdamW} \\
            Optimizer momentum & \multicolumn{5}{c}{$\beta_1$, $\beta_2$ = 0.9, 0.999} \\
            Batch size & 32 & 32 & 32 & 32 & 32 \\
            Training epoch & 100 & 10 & 10 & 5 & 10 \\
            \midrule
            Learning rate & 3e-5 & 1e-4 & 1e-4 & 1e-4 & 1e-4 \\
            Weight decay & \multicolumn{5}{c}{0}\\
            \bottomrule
        \end{tabular}
    }
    \label{hyper2}
\end{table}
%%\vspace{-\baselineskip}

Lastly, Table \ref{hyper3} details the default hyperparameter settings for the HyenaDNA model. %This model includes a few additional hyperparameters such as Embed dropout, Resid dropout, and Reverse complement augmentation, which are tailored for specific tasks like splice site prediction and enhancer activation prediction.
\begin{table}[h]
    \centering
    \caption{Default hyperparameter settings for HyenaDNA in downstream tasks}
    %%\vspace{-2mm}
    \resizebox{0.8\linewidth}{!}{
        \begin{tabular}{lcccc}
            \toprule
         & SSP & EMP & CPD\&PD & TF\\
        \midrule
        Optimizer & \multicolumn{4}{c}{AdamW} \\
        Optimizer momentum & \multicolumn{4}{c}{$\beta_1$, $\beta_2$ = 0.9, 0.999} \\
        Batch size & \multicolumn{4}{c}{256} \\
        Training epoch & \multicolumn{4}{c}{100} \\
        \midrule
        Learning rate & 6e-4 & 6e-4 & 7e-4 & 6e-4\\
        Weight decay & 0.20.0$^7$, 0.2$^8$ & 0.0$^{1,3,4}$, 0.1, 0.2$^5$ & 0.0 & 0.2\\
        Embed dropout & 0.1 & 0.0, 0.1$^{1,3,5}$, 0.2$^2$ & 0.0 & 0.2\\ 
        Resid dropout & 0.1$^7$, 0.2$^8$ & 0.0$^6$, 0.1, 0.2$^5$ & 0.1 & 0.1\\
        Reverse complement aug. & false & false & true & false\\
        \midrule
        \multicolumn{5}{l}{\scriptsize{$^1$H3, $^2$H3K4me1, $^3$H3K4me2, $^4$H3K36me3, $^5$H4, $^6$H4ac, $^7$splice site acceptor, $^8$splice site donor}} \\
        \bottomrule
        \end{tabular}
    }
    \label{hyper3}
\end{table}

%% file: TNNLS.bbl
\begin{thebibliography}{10}
\providecommand{\url}[1]{#1}
\csname url@samestyle\endcsname
\providecommand{\newblock}{\relax}
\providecommand{\bibinfo}[2]{#2}
\providecommand{\BIBentrySTDinterwordspacing}{\spaceskip=0pt\relax}
\providecommand{\BIBentryALTinterwordstretchfactor}{4}
\providecommand{\BIBentryALTinterwordspacing}{\spaceskip=\fontdimen2\font plus
\BIBentryALTinterwordstretchfactor\fontdimen3\font minus \fontdimen4\font\relax}
\providecommand{\BIBforeignlanguage}[2]{{%
\expandafter\ifx\csname l@#1\endcsname\relax
\typeout{** WARNING: IEEEtran.bst: No hyphenation pattern has been}%
\typeout{** loaded for the language `#1'. Using the pattern for}%
\typeout{** the default language instead.}%
\else
\language=\csname l@#1\endcsname
\fi
#2}}
\providecommand{\BIBdecl}{\relax}
\BIBdecl

\bibitem{devlin2018bert}
J.~Devlin, M.-W. Chang, K.~Lee, and K.~Toutanova, ``Bert: Pre-training of deep bidirectional transformers for language understanding,'' \emph{arXiv preprint arXiv:1810.04805}, 2018.

\bibitem{floridi2020gpt}
L.~Floridi and M.~Chiriatti, ``Gpt-3: Its nature, scope, limits, and consequences,'' \emph{Minds and Machines}, vol.~30, pp. 681--694, 2020.

\bibitem{zhou2021topicbert}
Y.~Zhou, L.~Liao, Y.~Gao, R.~Wang, and H.~Huang, ``Topicbert: A topic-enhanced neural language model fine-tuned for sentiment classification,'' \emph{IEEE Transactions on Neural Networks and Learning Systems}, vol.~34, no.~1, pp. 380--393, 2021.

\bibitem{zhao2023survey}
W.~X. Zhao, K.~Zhou, J.~Li, T.~Tang, X.~Wang, Y.~Hou, Y.~Min, B.~Zhang, J.~Zhang, Z.~Dong \emph{et~al.}, ``A survey of large language models,'' \emph{arXiv preprint arXiv:2303.18223}, 2023.

\bibitem{hua2023improving}
H.~Hua, X.~Li, D.~Dou, C.-Z. Xu, and J.~Luo, ``Improving pretrained language model fine-tuning with noise stability regularization,'' \emph{IEEE Transactions on Neural Networks and Learning Systems}, 2023.

\bibitem{khoury1983enhancer}
G.~Khoury and P.~Gruss, ``Enhancer elements,'' \emph{Cell}, vol.~33, no.~2, pp. 313--314, 1983.

\bibitem{riethoven2010regulatory}
J.-J.~M. Riethoven, ``Regulatory regions in dna: promoters, enhancers, silencers, and insulators,'' \emph{Computational biology of transcription factor binding}, pp. 33--42, 2010.

\bibitem{guo2022context}
Y.~Guo, D.~Zhou, P.~Li, C.~Li, and J.~Cao, ``Context-aware poly (a) signal prediction model via deep spatial--temporal neural networks,'' \emph{IEEE Transactions on Neural Networks and Learning Systems}, 2022.

\bibitem{wang2024grace}
J.-C. Wang, Y.-J. Chen, and Q.~Zou, ``Grace: Unveiling gene regulatory networks with causal mechanistic graph neural networks in single-cell rna-sequencing data,'' \emph{IEEE Transactions on Neural Networks and Learning Systems}, 2024.

\bibitem{gibbs2020human}
R.~A. Gibbs, ``The human genome project changed everything,'' \emph{Nature Reviews Genetics}, vol.~21, no.~10, pp. 575--576, 2020.

\bibitem{ji2021dnabert}
Y.~Ji, Z.~Zhou, H.~Liu, and R.~V. Davuluri, ``Dnabert: pre-trained bidirectional encoder representations from transformers model for dna-language in genome,'' \emph{Bioinformatics}, vol.~37, no.~15, pp. 2112--2120, 2021.

\bibitem{yang2022integrating}
M.~Yang, L.~Huang, H.~Huang, H.~Tang, N.~Zhang, H.~Yang, J.~Wu, and F.~Mu, ``Integrating convolution and self-attention improves language model of human genome for interpreting non-coding regions at base-resolution,'' \emph{Nucleic acids research}, vol.~50, no.~14, pp. e81--e81, 2022.

\bibitem{dalla2023nucleotide}
H.~Dalla-Torre, L.~Gonzalez, J.~Mendoza-Revilla, N.~L. Carranza, A.~H. Grzywaczewski, F.~Oteri, C.~Dallago, E.~Trop, H.~Sirelkhatim, G.~Richard \emph{et~al.}, ``The nucleotide transformer: Building and evaluating robust foundation models for human genomics,'' \emph{bioRxiv}, pp. 2023--01, 2023.

\bibitem{le2021transformer}
N.~Q.~K. Le, Q.-T. Ho, T.-T.-D. Nguyen, and Y.-Y. Ou, ``A transformer architecture based on bert and 2d convolutional neural network to identify dna enhancers from sequence information,'' \emph{Briefings in bioinformatics}, vol.~22, no.~5, p. bbab005, 2021.

\bibitem{nguyen2023hyenadna}
E.~Nguyen, M.~Poli, M.~Faizi, A.~Thomas, C.~Birch-Sykes, M.~Wornow, A.~Patel, C.~Rabideau, S.~Massaroli, Y.~Bengio \emph{et~al.}, ``Hyenadna: Long-range genomic sequence modeling at single nucleotide resolution,'' \emph{arXiv preprint arXiv:2306.15794}, 2023.

\bibitem{zhou2023dnabert}
Z.~Zhou, Y.~Ji, W.~Li, P.~Dutta, R.~Davuluri, and H.~Liu, ``Dnabert-2: Efficient foundation model and benchmark for multi-species genome,'' \emph{arXiv preprint arXiv:2306.15006}, 2023.

\bibitem{moore2013dna}
L.~D. Moore, T.~Le, and G.~Fan, ``Dna methylation and its basic function,'' \emph{Neuropsychopharmacology}, vol.~38, no.~1, pp. 23--38, 2013.

\bibitem{oubounyt2019deepromoter}
M.~Oubounyt, Z.~Louadi, H.~Tayara, and K.~T. Chong, ``Deepromoter: robust promoter predictor using deep learning,'' \emph{Frontiers in genetics}, vol.~10, p. 286, 2019.

\bibitem{kato2018sickle}
G.~J. Kato, F.~B. Piel, C.~D. Reid, M.~H. Gaston, K.~Ohene-Frempong, L.~Krishnamurti, W.~R. Smith, J.~A. Panepinto, D.~J. Weatherall, F.~F. Costa \emph{et~al.}, ``Sickle cell disease,'' \emph{Nature reviews Disease primers}, vol.~4, no.~1, pp. 1--22, 2018.

\bibitem{jawahar2019does}
G.~Jawahar, B.~Sagot, and D.~Seddah, ``What does bert learn about the structure of language?'' in \emph{ACL 2019-57th Annual Meeting of the Association for Computational Linguistics}, 2019.

\bibitem{de2022deepstarr}
B.~P. de~Almeida, F.~Reiter, M.~Pagani, and A.~Stark, ``Deepstarr predicts enhancer activity from dna sequence and enables the de novo design of synthetic enhancers,'' \emph{Nature Genetics}, vol.~54, no.~5, pp. 613--624, 2022.

\bibitem{van2008visualizing}
L.~Van~der Maaten and G.~Hinton, ``Visualizing data using t-sne.'' \emph{Journal of machine learning research}, vol.~9, no.~11, 2008.

\bibitem{fei2021optimizing}
H.~Fei, Y.~Zhang, Y.~Ren, and D.~Ji, ``Optimizing attention for sequence modeling via reinforcement learning,'' \emph{IEEE Transactions on Neural Networks and Learning Systems}, vol.~33, no.~8, pp. 3612--3621, 2021.

\bibitem{li2023bvit}
N.~Li, Y.~Chen, W.~Li, Z.~Ding, D.~Zhao, and S.~Nie, ``Bvit: Broad attention-based vision transformer,'' \emph{IEEE Transactions on Neural Networks and Learning Systems}, 2023.

\bibitem{bengio2009curriculum}
Y.~Bengio, J.~Louradour, R.~Collobert, and J.~Weston, ``Curriculum learning,'' in \emph{Proceedings of the 26th annual international conference on machine learning}, 2009, pp. 41--48.

\bibitem{pokholok2005genome}
D.~K. Pokholok, C.~T. Harbison, S.~Levine, M.~Cole, N.~M. Hannett, T.~I. Lee, G.~W. Bell, K.~Walker, P.~A. Rolfe, E.~Herbolsheimer \emph{et~al.}, ``Genome-wide map of nucleosome acetylation and methylation in yeast,'' \emph{Cell}, vol. 122, no.~4, pp. 517--527, 2005.

\bibitem{phaml2005qualitatively}
T.~H. Phaml, D.~H. Tran, T.~B. Ho, K.~Satou, and G.~Valiente, ``Qualitatively predicting acetylation and methylation areas in dna sequences,'' \emph{Genome Informatics}, vol.~16, no.~2, pp. 3--11, 2005.

\bibitem{wang2019splicefinder}
R.~Wang, Z.~Wang, J.~Wang, and S.~Li, ``Splicefinder: ab initio prediction of splice sites using convolutional neural network,'' \emph{BMC bioinformatics}, vol.~20, pp. 1--13, 2019.

\end{thebibliography}
